\documentclass{article}

\usepackage[preprint, nonatbib]{neurips_2026}
\usepackage[numbers]{natbib}

\usepackage[utf8]{inputenc} 
\usepackage[T1]{fontenc}    
\usepackage{hyperref}       
\usepackage{url}            
\usepackage{booktabs}       
\usepackage{amsfonts}       
\usepackage{nicefrac}       
\usepackage{microtype}      
\usepackage{xcolor}         
\usepackage{graphicx}
\usepackage{arydshln}
\usepackage{amsmath}
\usepackage{amssymb}
\usepackage{mathtools}
\usepackage{amsthm}
\usepackage{wrapfig}
\usepackage{titlesec}
\usepackage{multirow}
\usepackage{enumitem}
\usepackage{ifthen}
\usepackage{tcolorbox}
\tcbuselibrary{skins,breakable}

\definecolor{noteborder}{HTML}{8A6FBF}  
\definecolor{notefill}{HTML}{F3EEFB}    
\definecolor{notetitle}{HTML}{5E4B8B}   

\newcounter{finding}
\newcommand\finding[1]{%
    \refstepcounter{finding}%
    \begin{tcolorbox}[
        enhanced, breakable,
        colback=notefill, colframe=noteborder,
        boxrule=0pt, leftrule=2.5pt,
        arc=2pt, outer arc=2pt,
        left=6pt, right=6pt, top=3pt, bottom=3pt,
        before skip=2pt, after skip=1pt,
        fontupper=\small,
    ]%
    \textcolor{notetitle}{\textbf{\sffamily Finding~\thefinding.}}\ #1%
    \end{tcolorbox}%
}
\titlespacing*{\section}
{0pt}{1.0ex plus 0.3ex minus 0.2ex}{0.7ex plus 0.2ex}
\titlespacing*{\subsection}
{0pt}{0.8ex plus 0.2ex minus 0.2ex}{0.5ex plus 0.2ex}
\title{Studying Image Tokenizers as Visual Languages in Unified Multimodal Models}

\author{%
  Siting Li\textsuperscript{12}\thanks{Work done while interning at Amazon FAR.}\quad
  Zhengyang Wang\textsuperscript{2}\quad
  Simon Shaolei Du\textsuperscript{1}\quad
  Xi Chen\textsuperscript{2}\quad
  Yang Liu\textsuperscript{2}\quad\\[4pt]
  \normalfont
  \textsuperscript{1}University of Washington \qquad
  \textsuperscript{2}Amazon FAR \\[2pt]
}

\begin{document}

\maketitle

\begin{abstract}
  Image tokenizers define the ``visual language'' of unified multimodal models, yet are commonly studied through isolated metrics or generation-/understanding-only evaluations. These evaluations do not fully capture how visual tokens behave when modeled jointly with text.
  We build a controlled pure-autoregressive testbed and track task-specific validation losses during multimodal continual pretraining across text, image, text-to-image (T2I), and image-to-text (I2T) prediction. We examine how these losses scale and relate to downstream performance, then use them to study multimodal learnability---how well image and text tokens are jointly modeled---and tokenizer design.
  We find that (1) losses should be analyzed by task, since they exhibit distinct scaling behavior and rank tokenizers differently. (2) The loss--performance relationship depends on the predicted token space: for a fixed tokenizer, T2I and I2T losses correlate with generation quality, but across tokenizers, the T2I loss--performance relationship shifts with the image-token space, whereas I2T loss, computed over a shared text vocabulary, provides a more consistent signal. I2T loss also correlates with both generation and visual understanding performance after supervised finetuning.
  Using losses as a lens, we show that (3) better reconstruction does not necessarily yield lower task-specific losses or stronger downstream performance, and that (4) image tokenizer choice can affect text modeling under joint optimization. As case studies, we revisit three tokenizer design axes---the discriminator, semantic supervision, and vocabulary size---to examine their effects on joint modeling and downstream performance.
  Together, our testbed offers a complementary perspective on image tokenizers as visual languages, highlighting their interplay with text in joint multimodal training.
\end{abstract}
\section{Introduction}
\label{sec_intro}
\begin{figure}[ht]
  \begin{center}
    \centerline{\includegraphics[width=1.0\textwidth]{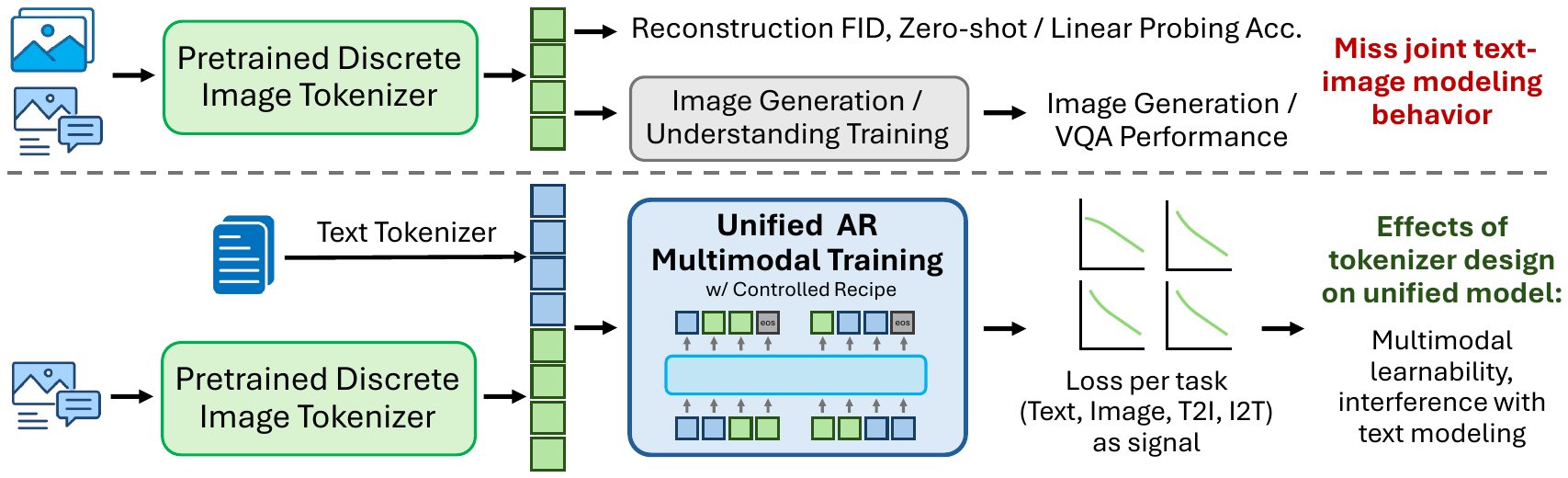}}
    \vskip -0.1in
    \caption{\textbf{Studying image tokenizers as visual languages in unified multimodal models.} Prior tokenizer studies focus on downstream-agnostic metrics (reconstruction/probing) or single-axis generation-/understanding-only pipelines, which miss joint text–image modeling behavior. We study tokenizers in the context of unified AR multimodal training, using task-specific losses as a signal to reveal how the tokenizer shapes downstream joint modeling.
    }
    \label{pipeline}
  \end{center}
  \vskip -0.45in
\end{figure}
Large-scale unified multimodal models have advanced rapidly in both visual generation and understanding by modeling images and text within a single framework. Among these, autoregressive (AR) modeling is a popular choice due to its compatibility with strong off-the-shelf language models~\cite{team2024chameleon, liu2024world, liquid, cui2025emu3}. In models that autoregressively predict both text and image tokens, a key component is the discrete image tokenizer, which converts raw pixels into discrete symbols that can be modeled alongside text~\cite{wu2024vila, shi2025scalable, ma2025unitok, lu2025atoken}. In this sense, an image tokenizer is not merely a preprocessing module; it defines the ``visual language'' that the language model must learn, align with text, and use for both visual generation and understanding.

However, existing tokenizer studies primarily examine this component outside the unified modeling context, relying on isolated reconstruction-based metrics (e.g., rFID~\cite{heusel2017gans}) or ImageNet classification accuracy~\cite{zhao2025qlip, zheng2025vision}, which do not necessarily translate into better downstream performance~\cite{yao2025reconstruction}. Others study tokenizers through generation-only or understanding-only downstream pipelines and report benchmark scores~\cite{tong2024cambrian1, yu2024image}. These end-to-end evaluations are valuable, but they still leave open how tokenizers affect joint modeling behavior. In unified pure-AR multimodal models, image and text tokens are modeled jointly under a shared model and training objective. Joint training with text may affect image-token modeling, while the choice of image tokenizer may in turn affect cross-modal alignment and text modeling. Whether these effects can be inferred from downstream-agnostic metrics or single-axis pipelines alone remains unclear. Consequently, tokenizer design choices, such as compression ratio and auxiliary losses, may be optimized without fully accounting for their effects on joint image--text modeling.

Motivated by this gap, we study \textbf{how the image tokenizer shapes multimodal modeling behavior in unified AR multimodal training}. In Section~\ref{sec_method}, we develop a controlled, pure-AR continual pretraining recipe using publicly available images and Qwen3 as the language backbone~\cite{yang2025qwen3}. Building upon this testbed, we use task-specific validation losses during continual pretraining to characterize tokenizer effects on text, image, text-to-image (T2I), and image-to-text (I2T) modeling. Applying this perspective requires addressing two questions. First, it is unclear whether task-specific losses exhibit similar scaling behavior during mixed image--text AR continual pretraining, or whether they must be interpreted separately. Second, it is unclear how these losses relate to downstream benchmark performance. Different image tokenizers define different prediction spaces, which may alter the loss--performance relationship. Moreover, visual benchmarks emphasize semantic faithfulness, perceptual quality, and cross-modal alignment, which are not directly measured by next-token prediction loss.

We therefore first examine how pretraining losses should be interpreted in unified multimodal training (Section~\ref{sec_loss_interpret}). Losses on text, image, T2I, and I2T all decrease with increasing data and model size, but exhibit distinct scaling behavior and rank tokenizers differently across tasks, making an averaged loss too coarse for this analysis. We then relate losses to generation and understanding benchmarks. When the tokenizer is fixed, T2I loss correlates with generation quality. I2T loss also tracks generation quality, suggesting that it captures aspects of multimodal training progress beyond caption prediction. Across tokenizers, however, the T2I loss--performance relationship shifts with the visual token space; vocabulary normalization reduces part of this shift, and the remaining gap is related to reconstruction fidelity. In contrast, I2T loss is computed over shared text tokens and remains a more consistent cross-tokenizer signal. I2T loss measured before supervised finetuning also correlates with post-finetuning generation and general VQA performance across tokenizers, supporting its use as a diagnostic signal.

Using task-specific losses as a lens, we make several observations about how the image tokenizer affects multimodal modeling behavior (Section~\ref{sec_case_study}). First, reconstruction fidelity can diverge from \emph{multimodal learnability}---how well image and text tokens are modeled under the shared AR objective---indicating that reconstruction metrics alone are insufficient for evaluating tokenizers in unified multimodal models. Second, tokenizer choice can affect text modeling through the image-token prediction objective under joint training, a cross-modal effect not captured by evaluating generation or understanding alone. We then revisit three tokenizer design axes through ablations and find that (1) a DINO-based discriminator improves reconstruction fidelity but does not consistently improve multimodal learnability or downstream performance, with VQAv2 performance declining; (2) semantic supervision encourages object-level image-token--word associations, improving multimodal learnability despite worse reconstruction fidelity; and (3) the relationship between vocabulary size and multimodal learnability is non-monotonic, though a larger vocabulary can still benefit downstream performance, potentially through higher reconstruction fidelity. Together, these results offer a complementary perspective on image tokenizers as visual languages, highlighting their interplay with text in joint multimodal training.

\section{Related Work}
\label{sec_related_work}
\textbf{Unified multimodal understanding and generation.} There have been recent advancements in building unified multimodal models, especially models for both visual understanding (captioning and VQA) and generation (text-to-image), in the hope of their mutual benefit. Various architectures have been proposed for unified models to jointly model text and image, including pure-autoregressive (AR) models (e.g., Chameleon \cite{team2024chameleon}, Emu3.5 \cite{cui2025emu3}, Liquid \cite{liquid}, Janus-Pro \cite{chen2025janus}, and LongCat-Next \cite{team2026longcat}), serial AR + diffusion models (e.g., BLIP3-o \cite{chen2025blip3} and MetaQuery \cite{pan2025transfer}), and hybrid AR + diffusion models (e.g., Transfusion \cite{zhou2024transfusion} and BAGEL \cite{deng2025emerging}). Among these models, some employ a unified architecture and tokenizer for understanding and generation (e.g., Chameleon and Liquid), while others use decoupled modules, Mixture-of-Experts (MoEs), or different inference modes for the two tasks, and some adopt separate image tokenizers, illustrating the difficulty in truly unifying the two tasks in both tokenization and downstream models \cite{zheng2025architecture}. We adopt a pure-AR setup to model both modalities through next-token prediction, providing a controlled setting for studying image tokenizer effects under a shared modeling objective.

\textbf{Loss-based analysis of unified multimodal models.} Loss-based metrics, such as Perplexity (PPL) or Bits-Per-Byte (BPB), are commonly tracked against compute to validate model efficiency: \citet{kaplan2020scaling} established held-out cross-entropy as a predictable scaling signal, \citet{magnusson2024paloma} adopt BPB because token-level perplexity is not comparable across tokenizers, and \citet{gadre2024scale} fit language-model loss directly to average downstream error. \citet{aghajanyan2023scaling} pioneered this analysis for mixed-modal models by establishing foundational scaling laws for mixed-modal loss, demonstrating that joint cross-entropy follows predictable power laws. Chameleon uses compute-loss curves to prove the stability of its early-fusion architecture at scale \cite{team2024chameleon}. More recently, Liquid \cite{liquid} utilizes these loss-vs-compute trends to demonstrate that inter-modality interference within a unified token space diminishes as model capacity increases. BAGEL \cite{deng2025emerging} and \citet{shukor2025scaling} identify emerging reasoning properties and the advantage of early-fusion and MoEs when scaling up compute, respectively. \citet{tong2026beyond} uses IsoFLOP analysis to uncover the scaling asymmetry between vision and language in unified models. The above works on multimodal training use loss to characterize scaling, optimization, or performance under a fixed token space; we instead ask how losses should be interpreted across image tokenizers and use losses to study image tokenizers' downstream effects.

\textbf{Studying image tokenizer properties.} (1) \textbf{Reconstruction.} PSNR and SSIM \cite{wang2004image} are pixel-wise metrics for image reconstruction quality, but are sensitive to noise and poorly align with human perception. Feature-based metrics, including rFID \cite{heusel2017gans}, IS \cite{salimans2016improved}, and LPIPS \cite{zhang2018unreasonable} on the ImageNet-1K validation set \cite{deng2009imagenet}, consider semantic and distributional quality of reconstructed images. Recent tokenizer benchmarks such as VTBench \cite{lin2025vtbench} and TokBench \cite{wu2025tokbench} measure how tokenizers preserve text, identity, and details, which are important for downstream tasks like OCR-VQA \cite{mishra2019ocr}. However, better reconstruction does not necessarily translate into better downstream performance, and could conflict with better generation \cite{yao2025reconstruction}. ETT \cite{wang2025ett} shares this motivation but tunes the tokenizer end-to-end with the downstream model, whereas we keep tokenizers frozen to compare fixed visual token spaces. (2) \textbf{Generation.} Tokenizers can be compared by training with the same generative model and measuring the generation quality by gFID on a fixed set \cite{wu2024vila, yu2024image, zhao2025qlip, lin2025toklip}, but generation-only evaluation does not establish how tokenizer choice affects both generation and understanding under joint training. (3) \textbf{Semantics for understanding.} Zero-shot accuracy and linear probing accuracy on ImageNet-1K have been used for both continuous and discrete tokenizers \cite{radford2021learning, oquab2023dinov2, zheng2025vision, zhao2025qlip}, but higher classification accuracy does not necessarily translate into better downstream visual understanding, as observed in UniTok \cite{ma2025unitok}. GigaTok \cite{xiong2025gigatok} suggests AR probing accuracy as a better proxy to predict downstream performance when training with a larger AR model. TA-Tok \cite{han2025vision} directly measures downstream performance at varying data scales. Cambrian-1 \cite{tong2024cambrian1} pioneered leveraging multimodal large language models as an interface for tokenizer evaluation, but is restricted to continuous tokenizers. Apart from the above, (4) \textbf{codebook usage} is often reported in earlier work, but recent tokenizers can achieve near 100\% usage with techniques like entropy loss even when they have a large vocabulary \cite{zhu2024scaling}.

\section{Preliminaries}
\label{sec_tokenizer}
\setlength{\tabcolsep}{1pt}
In this section, we review discrete image tokenizers and the key design axes along which they differ: model and codebook architecture, bitwise compression ratio, and training objectives. Table~\ref{tab5_tokenizer} summarizes the tokenizers used in our study.

The VQGAN-based image tokenizers studied here~\cite{esser2021taming, yu2021vector} follow an encoder-quantizer-decoder architecture. The encoder maps an input RGB image $x\in \mathbb{R}^{H\times W\times 3}$ to a sequence of $K$ continuous vectors $f\in \mathbb{R}^{K\times D}$, where $K$ is the number of image tokens and $D$ is the feature dimension. The quantizer maintains a codebook $Z\in \mathbb{R}^{B\times D}$ of vocabulary size $B$ and replaces each encoder vector with its nearest codebook entry, yielding quantized vectors $z\in \mathbb{R}^{K\times D}$. The corresponding codebook indices form the discrete image-token sequence modeled by the autoregressive model. The decoder reconstructs an RGB image $\hat{x}\in \mathbb{R}^{H\times W\times 3}$ from $z$. A standard VQGAN is trained with a vector-quantization (VQ) loss and an autoencoding (AE) loss:
\begin{align}
    \mathcal{L}_{VQ}&=||\text{sg}(f)-z||_2^2 + \beta||f-\text{sg}(z)||_2^2, ~~\mathcal{L}_{AE}=\mathcal{L}_2(x,\hat{x})+\mathcal{L_P}(x,\hat{x})+\mathcal{L_G}(x,\hat{x}), \label{formula}
\end{align}
where $\text{sg}(\cdot)$ denotes the stop-gradient operation, $\beta$ weights the commitment loss, and $\mathcal{L}_2, \mathcal{L_P}, \mathcal{L_G}$ are the $L_2$ reconstruction loss, the LPIPS perceptual loss~\cite{zhang2018unreasonable}, and the adversarial (GAN) loss from the discriminator, respectively.

\begin{wraptable}[15]{r}{0.49\textwidth}
\vspace{-22pt}
  \begin{center}
    \caption{\textbf{Tokenizers studied in our framework:} IBQ~\cite{shi2025scalable}, GigaTok~\cite{xiong2025gigatok}, and UniTok~\cite{ma2025unitok}. $B$: vocabulary size; $\mathcal{L}_{sem}$: semantic loss; Disc.: discriminator type; Usage: codebook usage; rFID: reconstruction FID on ImageNet-1K; $\dagger$: models trained by us.}
   \label{tab5_tokenizer}
   \vskip -0.0in
   {\footnotesize
  \begin{tabular}{lcccccccc}
  \toprule
    Tokenizer & Size &\textbf{$B$} &$\mathcal{L}_{sem}$& Disc. &Usage & rFID$\downarrow$ \\
    \midrule
    IBQ-1024 & 0.1B &  1024&  & PatchGAN& 99\% &2.24\\
    IBQ-8192  & 0.1B &  8192 &  &PatchGAN &98\% &1.87\\
    IBQ-16384  & 0.1B &  16384&  & PatchGAN&96\% & 1.37\\ \hdashline \noalign{\vskip 1pt}
    GigaTok  &  0.6B&  16384& \checkmark&PatchGAN&100\% & 0.81\\
    GigaTok-DINO  & 0.6B&  16384 & \checkmark& DINO & 100\% &0.51 \\
    \hdashline \noalign{\vskip 1pt}
    UniTok$^\dagger$  &  0.8B&  16384& &DINOv2-S& 100\% & 1.86\\
    UniTok-sem$^\dagger$  & 0.8B&  16384 &  \checkmark& DINOv2-S &100\% &2.23 \\
    \bottomrule
  \end{tabular}}
  \end{center}
  \vspace{-10pt}
\end{wraptable}

\textbf{Architecture.} On the encoder/decoder choices, prior work adopts various architectures, including ConvNet~\cite{esser2021taming, sun2024autoregressive}, ViT~\cite{yu2021vector}, and ViT-based models with learnable latent queries~\cite{yu2024image}. On the codebook design, some tokenizers use multiple sub-codebooks~\cite{qu2025tokenflow, song2025dualtoken, ma2025unitok} to represent each spatial position with multiple indices. Modeling these indices introduces additional choices in sequence organization and prediction architecture, complicating controlled comparisons. We therefore focus on single-codebook tokenizers throughout this work.

\textbf{Bitwise compression ratio.} The bitwise compression ratio is determined by the input image resolution $H\times W$, the number of image tokens $K$, and the vocabulary size $B$. In this work, we fix the number of image tokens to $K=16\times 16$ and the input resolution to $H\times W=256\times 256$, and study the \textbf{vocabulary size $B$} as the main compression-related axis using the IBQ tokenizer family~\cite{shi2025scalable} (Section~\ref{sec_vocabulary_size}). We defer to future work the study of varying $H\times W$ (e.g., any-resolution image tokenizers~\cite{lu2025atoken}) and varying $K$ (e.g., highly compressed 1D tokenizers~\cite{bachmann2025flextok, kim2025democratizing}), as changing image resolution alters the visual detail available to the model, while changing $K$ alters sequence length and the image-token budget during training, introducing additional factors into the comparison.

\textbf{Training objectives.} Beyond the standard losses in Eq.~(\ref{formula}), we study two design choices in the training objective. On the supervision signal, recent tokenizers add \textbf{semantic supervision}, such as a CLIP contrastive loss~\cite{ma2025unitok}, to encourage the tokens to capture high-level semantics; we study its effect on downstream joint modeling by training single-codebook UniTok variants without and with this loss, denoted UniTok and UniTok-sem, respectively~\cite{ma2025unitok}, in Section~\ref{sec_semantic_loss}. On the adversarial loss $\mathcal{L}_G$, some tokenizers replace the standard PatchGAN discriminator~\cite{isola2017image} with a \textbf{DINO-based discriminator}~\cite{caron2021emerging} to obtain lower rFID~\cite{tian2024visual, li2024imagefolder}; we revisit whether this reconstruction gain carries over to downstream joint modeling using GigaTok with the two discriminators (denoted GigaTok and GigaTok-DINO)~\cite{xiong2025gigatok} in Section~\ref{sec_discriminator}.

\section{Framework Construction}
\label{sec_method}
In this section, we describe the framework for exploring the image tokenizer's effect on downstream unified multimodal training. In Section~\ref{sec_recipe}, we introduce the training recipe, which extends a pretrained text language model into a unified autoregressive multimodal model through continual pretraining and supervised finetuning. In Section~\ref{sec_loss}, we define the validation loss we use to measure modeling quality across tasks. Further training details and tokenizer information are provided in Appendix~\ref{appendix_training}.

\subsection{Training Recipe}
\label{sec_recipe}
\textbf{Overall setup.} We extend pretrained Qwen3 language models~\cite{yang2025qwen3} into unified autoregressive multimodal models by expanding their vocabularies to include discrete image tokens. Our main experiments use three dense model sizes: Qwen3-0.6B, 1.7B, and 4B. Given an image tokenizer with vocabulary size $B$, we add $B$ new learnable embeddings to the base model's vocabulary, initialized from a multivariate normal distribution matching the original embeddings' mean and covariance. We also extend the LM head to predict the newly added image tokens. We add special tokens ⟨boi⟩ and ⟨eoi⟩ to mark the start and end of image-token sequences. During training, we remove the text conditioning from $10\%$ of T2I samples and use ⟨unconditional⟩ to mark unconditional image generation, enabling classifier-free guidance (CFG) at evaluation~\cite{ho2022classifier}. The resulting model accepts and produces both text tokens and flattened image-token IDs. We continually pretrain it on mixed-modal data and then apply supervised finetuning (SFT) on multimodal instruction-following data, as detailed below.

\textbf{Continual pretraining stage.} In this stage, we train the unified model on large-scale image-text and pure-text data.

(1) \textbf{Data mixture and preprocessing.} Our largest data scale is 60M samples, comprising 6.6M pure-text samples from DataComp-LM~\cite{li2024datacomp} and 53.3M image-text samples from three sources: (a) LAION-Aesthetics~\cite{schuhmann2022laion}, filtered by aesthetic score $\geq 5.5$ and recaptioned by InternVL3-1B~\cite{zhu2025internvl3}; (b) JourneyDB~\cite{sun2023journeydb}, recaptioned by GPT-3.5; and (c) BLIP3o-Pretrain-Short-Caption~\cite{chen2025blip3}. We resize all images to $256\times 256$ and pre-tokenize them before training.

(2) \textbf{Loss and token sequence formatting.} For pure-text samples, we adopt the standard cross-entropy loss for next-token prediction in training. For image-text samples, we initially assign $80\%$ to text-to-image (T2I) prediction and $20\%$ to image-to-text (I2T) prediction. As described above, $10\%$ of the T2I samples are converted to unconditional image generation. The conditional sequence formats are:
\begin{align*}
    &\text{T2I: \{text\}\{prompt\}⟨boi⟩\{\textbf{image tokens}\}⟨\textbf{eoi}⟩⟨\textbf{eos}⟩},\\
    &\text{I2T: ⟨boi⟩\{image tokens\}⟨eoi⟩\{prompt\}\{\textbf{text}\}⟨\textbf{eos}⟩}.
\end{align*}
Following Liquid~\cite{liquid}, we compute the cross-entropy loss only on the \textbf{bold} tokens (conditional next-token prediction), leaving the prompt and the conditioning modality unscored. The prompts are listed in Appendix~\ref{appendix_training_prompt}.

(3) \textbf{Hyperparameters.} We use a Warmup-Stable-Decay (WSD) schedule with a $0.03$ warmup ratio and linear decay over the last $20\%$ of steps. For the 0.6B model, we sweep the learning rate (lr) over $\{3\text{e-}5, 1\text{e-}4\}$ and the batch size (bs) over $\{512, 1024, 2048\}$, forming a $2\times 3$ grid. Because different tasks favor different hyperparameter settings (Appendix~\ref{appendix_hyperparameters}), we adopt a balanced setting of (lr $=3\text{e-}5$, bs $=512$) for the main experiments and reuse it for the 1.7B and 4B models. We also experimented with larger learning rates ($3\text{e-}4$ for 0.6B and $1\text{e-}4$ for 4B), but found that they caused unstable training with loss spikes.

\textbf{Supervised finetuning (SFT) stage.} We finetune the models for 2 epochs on 4.9M instruction-following samples. The mixture combines 1M LMSYS-Chat pure-text instructions~\cite{zheng2023lmsys}, 2.9M multimodal instruction and captioning samples from Mini-Gemini~\cite{li2025mini}, and 1M text-to-image samples (LAION-Aesthetics sampled from the continual-pretraining distribution, together with JourneyDB and BLIP3o-60K data~\cite{chen2025blip3}). We use a cosine schedule with a $0.03$ warmup ratio, a peak learning rate of $5\text{e-}5$, and batch size $1024$.

Further details about data of both stages, including the subset proportions, can be found in Appendix~\ref{appendix_training_proportion}. All images used in the training are publicly accessible.

\setlength{\tabcolsep}{1pt}
\begin{wraptable}[9]{r}{0.49\textwidth}
\vspace{-24pt}
  \begin{center}
    \caption{\textbf{Recipe validation.} Our 8B model obtains similar GenAI-Bench \cite{li2024genai}, WISE \cite{niu2025wise}, and VQA (Mean) scores to Liquid-7B~\cite{liquid}, while trailing on MJHQ-30K \cite{li2024playground}. ``Data'' refers to the number of samples seen in continual pretraining. }
   \label{tab1_performance}
   \vskip -0.1in
   {\footnotesize
  \begin{tabular}{lcccccccc}
  \toprule
    Model & Data &GenAI$\uparrow$ & MJHQ-30K$\downarrow$ & WISE$\uparrow$ & VQA (Mean)$\uparrow$ \\
    \midrule
    Liquid-7B & 90M &0.72 & 5.47 & 0.41 & 61.40 \\
    Ours (8B) & 60M & 0.73 &  10.55& 0.38 & 60.83\\
    \bottomrule
  \end{tabular}}
  \end{center}
\vspace{-16pt}
\end{wraptable}
\textbf{Training recipe validation.} We verify our training recipe by training a Qwen3-8B model with the Chameleon tokenizer (Table~\ref{tab1_performance}). Compared with Liquid-7B, our model reaches similar performance on GenAI-Bench, WISE, and VQA (averaged over VQAv2, GQA, TextVQA, and POPE), while trailing on MJHQ-30K. Full benchmark scores and image generation examples are provided in Appendix~\ref{appendix_training_comparison}. We view this experiment as a validation that the recipe serves as a reliable testbed for studying tokenizers, rather than as evidence of achieving superior performance.

\subsection{Validation Loss}
\label{sec_loss}
We use validation loss, defined as the mean negative log-likelihood over supervised tokens, to measure how well a model fits held-out data, following common practice in language-model scaling studies~\cite{kaplan2020scaling, hoffmann2022training, aghajanyan2023scaling}. Let $N$ be the evaluation set. For a sequence $t\in N$, let $S(t)\subseteq\{1,\dots,|t|\}$ denote its \emph{supervised} positions: the tokens on which the loss is computed (the \textbf{bold} spans of the T2I/I2T formats in Section~\ref{sec_recipe}; all positions for pure-text samples). The model's log-likelihood on the supervised tokens is
\begin{align}
    l = \sum_{t\in N}\sum_{i\in S(t)} \ln p(t_i\mid t_{<i}),
    \label{eq_loglik}
\end{align}
where conditioning tokens (e.g., the text prompt or the conditioning image) are never scored but still enter through the context $t_{<i}$. Using $T(N)=\sum_{t\in N}|S(t)|$ for the number of supervised tokens, we report the mean per-token loss
\begin{align}
    \mathcal{L} = -\frac{l}{T(N)}.
    \label{eq_loss}
\end{align}

\textbf{Note on loss interpretation.} Unlike the bits-per-byte loss common in language modeling, which upper-bounds a description length of the raw data~\cite{gao2020pile, magnusson2024paloma}, our validation loss on image tokens is computed over the tokenizer's discrete \emph{latent} codes. Because image tokenization is lossy, the model's likelihood on these codes cannot be converted into a description length for the original pixels. We therefore interpret $\mathcal{L}$ as a measure of \textbf{latent-modeling} quality for a fixed visual language. Comparisons across image tokenizers require accounting for differences in their token spaces, as examined in Section~\ref{sec_loss_performance}.

\textbf{Evaluation protocol.} We hold out $50{,}000$ pure-text and $50{,}000$ image-text samples, drawn from the training distribution, as the evaluation set. This yields validation losses for four tasks: \textbf{text}, unconditional \textbf{image}, \textbf{text-to-image (T2I)}, and \textbf{image-to-text (I2T)}. We apply the same loss-masking rules as in training: conditioning tokens provide context but do not contribute directly to the loss. Empirically, losses over different image sources behave similarly, so we report LAION-Aesthetics as our main proxy, denoted LAION-(Image, T2I, I2T); losses on the other sources are given in Appendix~\ref{appendix_image_loss}. In loss-scaling plots, the $x$-axis FLOPs denote the compute used in continual mixed-modal pretraining.

\section{Interpreting Loss in Unified Multimodal Training}
\label{sec_loss_interpret}
Before using pretraining loss to study tokenizers, we first examine how it should be interpreted in unified multimodal training. We begin by asking whether a single unified loss metric can characterize tokenizer behavior in downstream modeling, or whether the task-specific losses must be analyzed separately (Section~\ref{sec_compute_loss}). We then examine whether these losses align with the performance on image generation benchmarks and how they can be calibrated across tokenizers (Section~\ref{sec_pretraining}). Finally, we test how they relate to post-SFT performance (Section~\ref{sec_post_training}).
\subsection{Task-Specific Loss Scaling}
\label{sec_compute_loss}
\begin{figure}[t]
  \begin{center}
    \centerline{\includegraphics[width=1.0\textwidth]{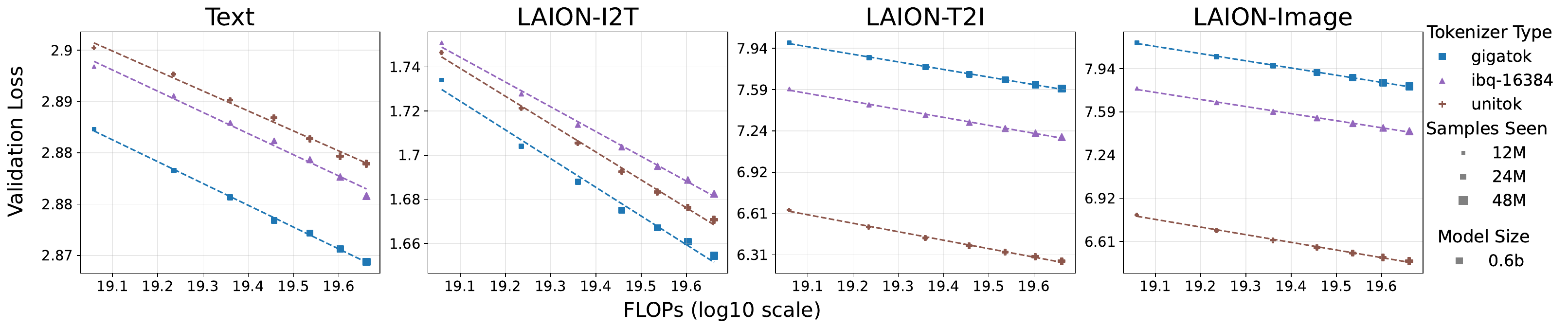}}
    \vskip -0.15in
    \caption{
      \textbf{Validation losses scale with data size differently across tasks} for the 0.6B model. The y-axis is in log scale. The lines are linear fits to the dots per tokenizer.
    }
    \label{dataflops-loss}
  \end{center}
  \vskip -0.35in
\end{figure}
\begin{figure}[t]
  \begin{center}
    \centerline{\includegraphics[width=1.0\textwidth]{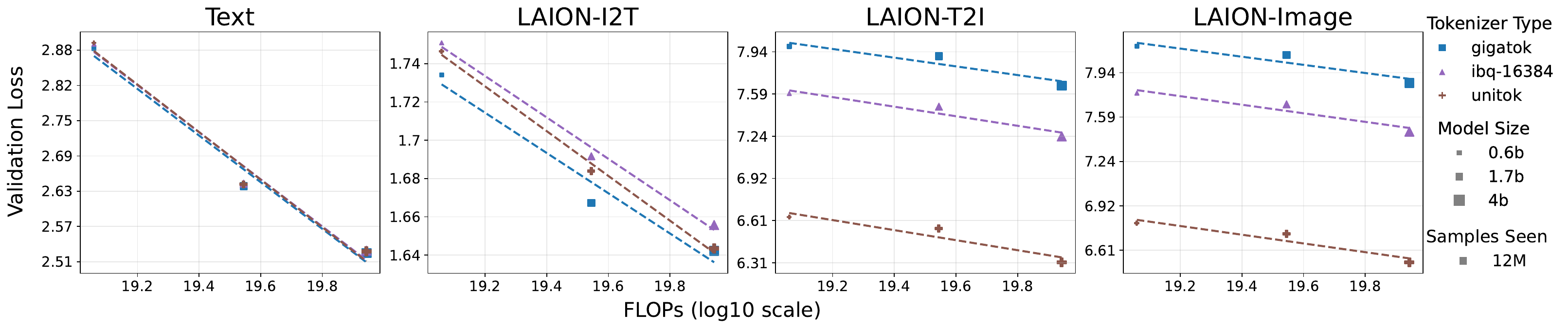}}
    \vskip -0.15in
    \caption{\textbf{Validation losses scale with model size differently across tasks} on 12M data. The y-axis is in log scale. The lines are linear fits to the dots per tokenizer.
    }
    \label{modelflops-loss}
  \end{center}
  \vskip -0.4in
\end{figure}
Our setting differs from standard language-only pretraining in three aspects: we continually train from a text-only pretrained model; the training is mixed-modal, spanning four distinct tasks; and the T2I and I2T losses are conditional. These differences make the loss harder to interpret: it is not obvious whether the four task-specific losses share a common scaling behavior or must be analyzed separately. We therefore first examine how each task's loss scales with data and model size, and then ask whether any single signal derived from the loss yields a tokenizer ranking that is consistent across tasks.

\textbf{All task-specific losses scale with data and model size.} We plot validation loss against pretraining FLOPs for three tokenizers, varying data scale (Figure~\ref{dataflops-loss}) and model size (Figure~\ref{modelflops-loss}). Both figures show that the losses on all four tasks decrease smoothly and roughly follow a power law (a linear trend in the log-log plots) under both data and model scaling. For the data-scaling analysis, we use intermediate, pre-annealing checkpoints from a single training run for each tokenizer--model-size configuration, with lr $=3\text{e-}5$ and bs $=512$. Annealed checkpoints follow the same power-law trends, with annealing mainly reducing the text loss and leaving the image-side losses largely unchanged (Appendix~\ref{appendix_annealed_loss}). Results across hyperparameters are deferred to Appendix~\ref{appendix_hyperparameters}, and further continual-pretraining dynamics to Appendix~\ref{appendix_training_dynamics}.

\textbf{The scaling patterns differ across tasks.} Although all four losses scale, they do so in qualitatively different ways, which makes an aggregated loss too coarse to serve as an analysis signal. \emph{(1) Text loss is largely governed by language-model initialization, while image-related losses are shaped more by multimodal continual pretraining.} Under data scaling at a fixed model size (Figure~\ref{dataflops-loss}), text loss changes only slightly, whereas image, T2I, and I2T losses decrease more substantially. This suggests that text-modeling quality is largely inherited from the pretrained language backbone, while image-related tasks benefit more from multimodal continual pretraining. Under model scaling (Figure~\ref{modelflops-loss}), text loss also decreases, consistent with initializing larger models from stronger pretrained language backbones; this trend therefore cannot be attributed to multimodal continual pretraining alone~\cite{zhang2024pre}. \emph{(2) Image-token modeling drives T2I, while I2T reflects cross-modal alignment.} The T2I loss closely tracks the unconditional image loss, indicating that image-token prediction accounts for much of the difficulty in T2I. The I2T loss, though also computed over text tokens, behaves differently from the text loss, indicating that conditioning on image tokens introduces a distinct cross-modal signal.

\textbf{Tokenizer rankings differ across tasks under both data and model scaling.} No single ordering of the three tokenizers holds across tasks. In Figure~\ref{dataflops-loss}, UniTok has the highest text loss yet the lowest T2I loss, while GigaTok has the lowest text-related losses but the highest T2I loss. Consequently, no single task loss yields a tokenizer ranking that holds across tasks, and a simple average can obscure these task-dependent differences. These task-dependent rankings suggest trade-offs between text modeling and image-token prediction under the unified AR objective. Stronger semantic alignment, in turn, can help I2T even when a tokenizer is not optimal for pure-text modeling, as seen in the comparison between UniTok and IBQ-16384 in Figure~\ref{dataflops-loss}.

\finding{Unified multimodal training loss should be analyzed per task: task-specific losses exhibit distinct scaling behavior that an averaged loss obscures, and no single tokenizer ranking holds across tasks.}

\subsection{Loss--Performance Relation}
\label{sec_loss_performance}
We next study how pretraining losses relate to benchmark performance. These validation losses measure next-token modeling quality, whereas downstream benchmarks emphasize semantic faithfulness, perceptual quality, and cross-modal alignment. We therefore ask the following questions:
\begin{itemize}[noitemsep, left=0pt]
\item (Section~\ref{sec_pretraining}) In pretraining, do the losses align with image-generation benchmark scores, and are these relationships consistent across tokenizers?
\item (Section~\ref{sec_post_training}) Do the losses serve as signals of post-SFT benchmark performance?
\end{itemize}
\textbf{Benchmarks.}\label{sec_benchmarks} We evaluate the continually pretrained checkpoints under varying hyperparameters on GenAI-Bench~\cite{li2024genai} and MJHQ-30K~\cite{li2024playground}, scored by VQAScore ($\uparrow$) and $\log_{10}$ gFID ($\downarrow$), respectively. We then apply our SFT recipe to the pretraining checkpoints after annealing and re-evaluate generation performance, together with general visual understanding on VQAv2~\cite{goyal2017making} and GQA~\cite{hudson2019gqa}. Qualitative examples on GenAI-Bench and further benchmark details are in Appendix~\ref{appendix_generation_benchmarks}.
\subsubsection{Alignment between Losses and T2I Quality in Pretraining}
\label{sec_pretraining}
\begin{figure}[t]
  \vskip 0.0in
  \begin{center}
    \centerline{\includegraphics[width=1.0\textwidth]{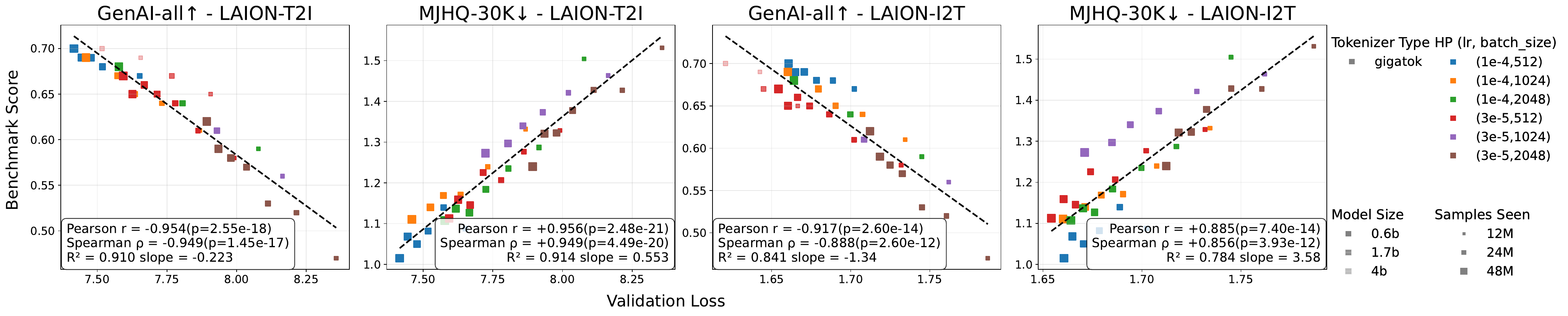}}
    \vskip -0.15in
    \caption{
      \textbf{Both T2I loss and I2T loss align with text-to-image generation quality when tokenizer type is controlled.} The lines are linear fits to all dots. 
    }
    \label{loss-genai-gigatok}
  \end{center}
  \vskip -0.35in
\end{figure}
\begin{figure}[t]
  \vskip -0.0in
  \begin{center}
    \centerline{\includegraphics[width=1.0\textwidth]{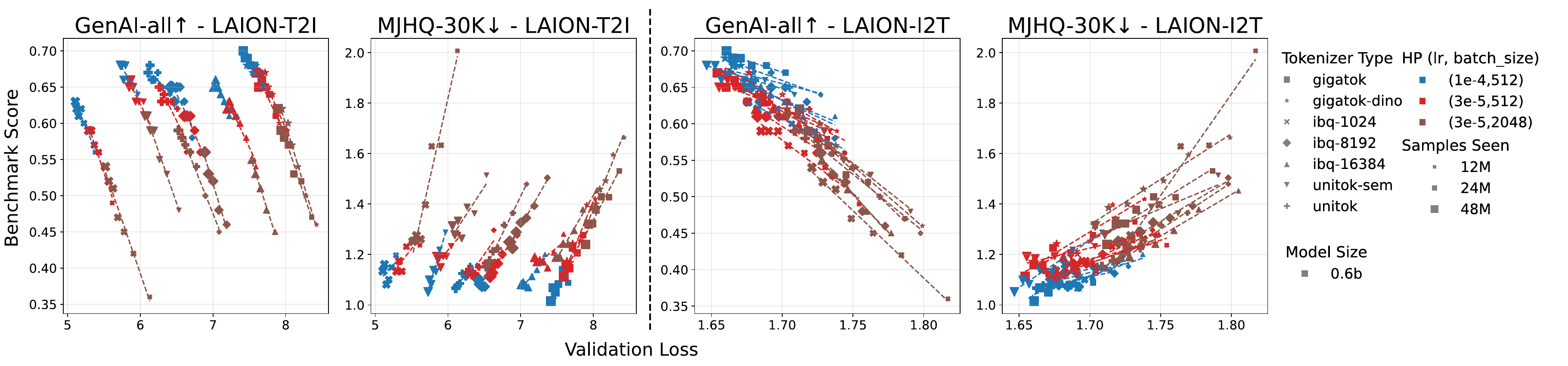}}
    \vskip -0.15in
    \caption{
      \textbf{T2I loss does not align with text-to-image generation quality across tokenizers} (the first two plots), \textbf{while I2T loss aligns with generation quality} (the last two plots). The lines are linear fits to the dots per tokenizer and per hyperparameter.
    }
    \label{loss-genai}
  \end{center}
  \vskip -0.45in
\end{figure}

\begin{figure}[t]
  \centering
  \begin{minipage}[t]{0.575\textwidth}
    \centering
    \includegraphics[width=\linewidth]{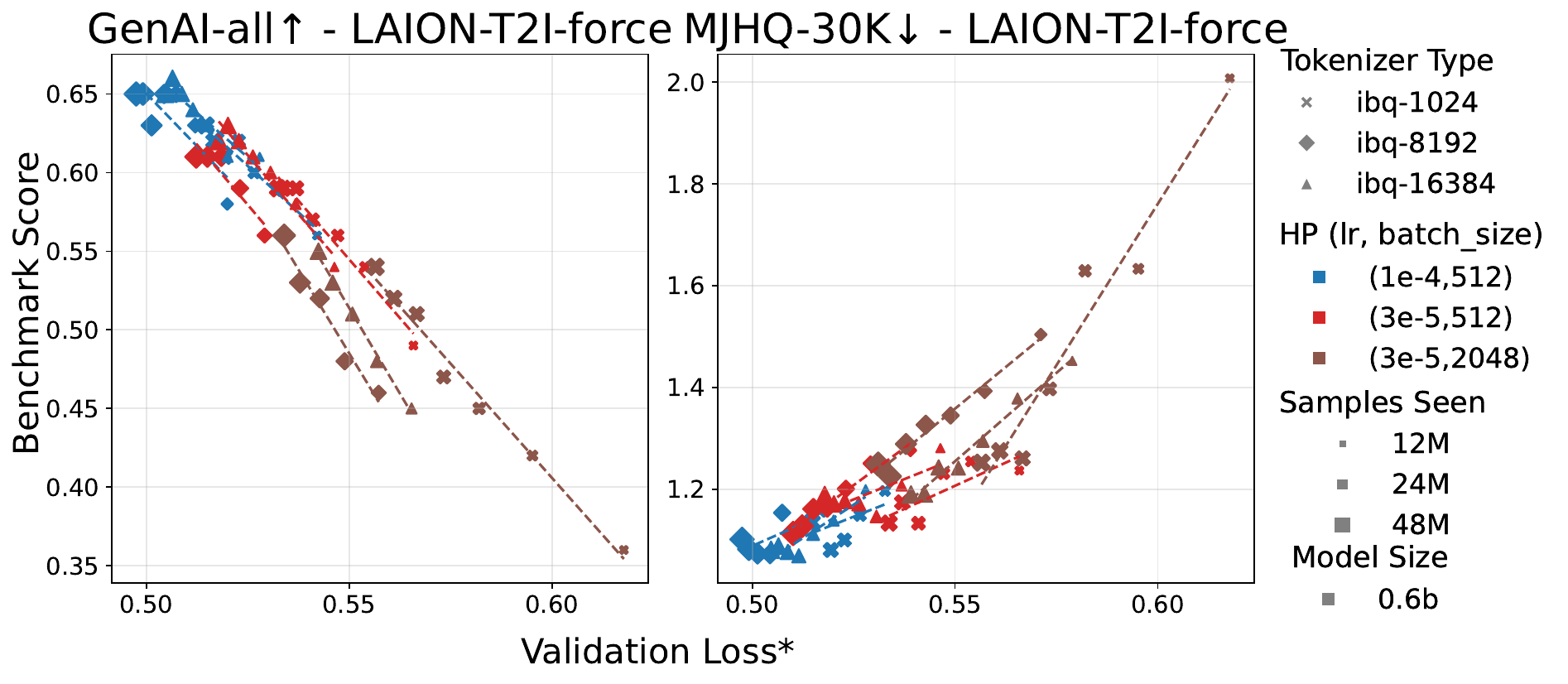}
    \vskip -0.15in
    \caption{\textbf{After vocabulary normalization, a more consistent T2I loss--performance relation exists across IBQ tokenizers.} The lines are linear fits to the dots per hyperparameter. Quantitative results are in Table~\ref{appendix_tab_norm_compare}.}
    \label{loss-genai-ibq-B}
  \end{minipage}\hfill
  \begin{minipage}[t]{0.405\textwidth}
    \centering
    \includegraphics[width=\linewidth]{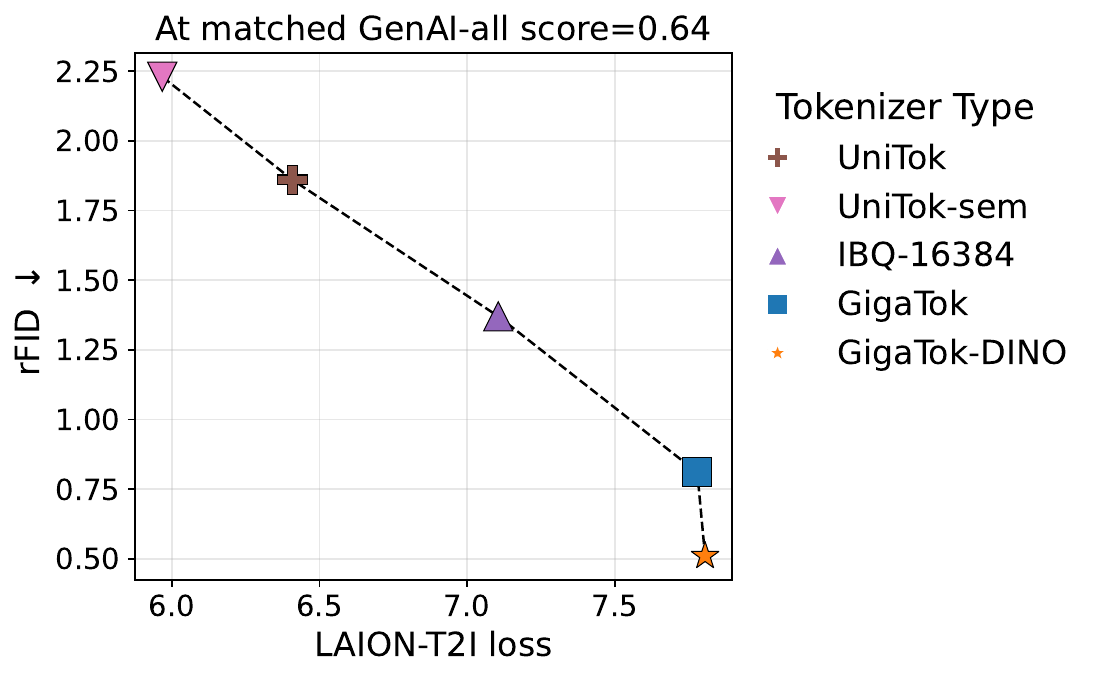}
    \vskip -0.15in
    \caption{\textbf{The T2I loss--performance relation shift is related to differences in reconstruction fidelity.} The rFID is negatively correlated with T2I loss when the GenAI benchmark score is controlled.}
    \label{rfid-loss}
  \end{minipage}
  \vskip -0.15in
\end{figure}
We study the alignment between pretraining losses and T2I generation quality, which can be evaluated directly on the pretrained checkpoints without SFT.

\textbf{Within a tokenizer, both T2I and I2T loss align with T2I quality.} Since T2I generation predicts image tokens conditioned on text, one might expect the T2I validation loss to be the most relevant pretraining signal for generation quality. This holds when the tokenizer is fixed: as shown in Figure~\ref{loss-genai-gigatok}, T2I loss aligns strongly with GenAI-Bench VQAScore and MJHQ-30K gFID across data scales and hyperparameter settings. More surprisingly, I2T loss also correlates with generation quality, albeit more weakly, suggesting that it captures aspects of multimodal training progress beyond caption prediction.

\textbf{Across tokenizers, the T2I loss--performance relation shifts with the image-token space.} We next ask whether both alignments hold across tokenizers. I2T loss is computed over a shared text vocabulary, providing a common basis for comparison. Empirically, Figure~\ref{loss-genai} shows a more consistent relationship between I2T loss and generation quality across tokenizers. T2I loss, however, exhibits tokenizer-dependent shifts in its relationship with generation quality. Thus, T2I loss remains a useful modeling diagnostic, but the same loss need not correspond to the same image quality across tokenizers, since each defines a different image-token space.

\textbf{Vocabulary normalization reduces the cross-tokenizer shift in the T2I loss--performance relation.} We hypothesize that vocabulary size contributes to the numerical scale of T2I loss. A natural reference is $\log_2 B$, the entropy of a uniform distribution over $B$ image tokens. We therefore define the \emph{vocabulary-normalized} image loss as $\mathcal{L}^{\ast} = \mathcal{L} / \log_2 B$, where $B$ denotes the image vocabulary size. Evaluated on IBQ tokenizers with $B\in\{1024,8192,16384\}$, $\mathcal{L}^{\ast}$ yields a more consistent T2I loss--performance relation across the IBQ variants (Figure~\ref{loss-genai-ibq-B}). Further details are in Appendix~\ref{appendix_loss_normalization}.

\textbf{The residual cross-tokenizer gap is linked to reconstruction fidelity.} Among tokenizers with the same vocabulary size ($16384$), Figure~\ref{rfid-loss} shows that, at a fixed benchmark score, T2I loss follows the reverse order of the rFID ranking: a tokenizer with worse reconstruction fidelity reaches the same generation quality at a lower T2I loss. This association suggests that reconstruction fidelity may help explain the residual differences in the T2I loss--performance relationship at a fixed vocabulary size. These results motivate considering reconstruction fidelity alongside T2I loss when comparing generation performance across tokenizers.
\par 
\finding{Token space shapes the loss--performance relationship. I2T loss, computed over a shared text vocabulary, provides a more consistent cross-tokenizer signal of generation performance, whereas T2I loss exhibits tokenizer-dependent shifts. Vocabulary normalization reduces these shifts within the IBQ family, while residual differences at a fixed vocabulary size are associated with reconstruction fidelity.}
\subsubsection{Correlation between Losses and Post-SFT Performance}
\label{sec_post_training}
We ask whether pretraining losses remain a useful signal of benchmark performance after supervised finetuning, which introduces additional variation beyond pretraining.
\label{genai_post}

\begin{figure}[t]
  \vskip 0.0in
  \begin{center}
    \centerline{\includegraphics[width=1.0\textwidth]{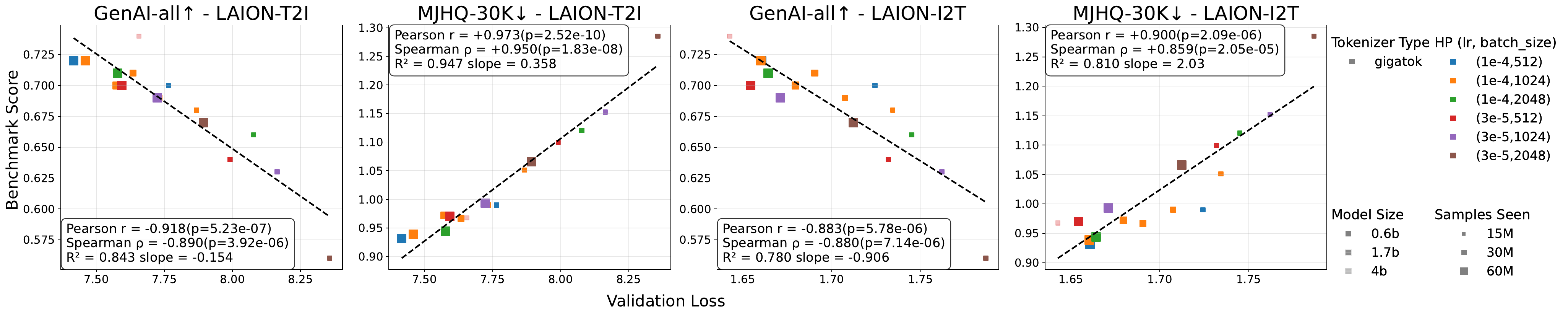}}
    \vskip -0.15in
    \caption{\textbf{Both I2T and T2I loss show strong correlation with post-SFT generation performance when tokenizer type is controlled.} We use the checkpoints after annealing for loss evaluation.}
    \label{loss-genai-post}
  \end{center}
  \vskip -0.35in
\end{figure}
\begin{figure}[t]
  \vskip 0.0in
  \begin{center}
    \centerline{\includegraphics[width=1.0\textwidth]{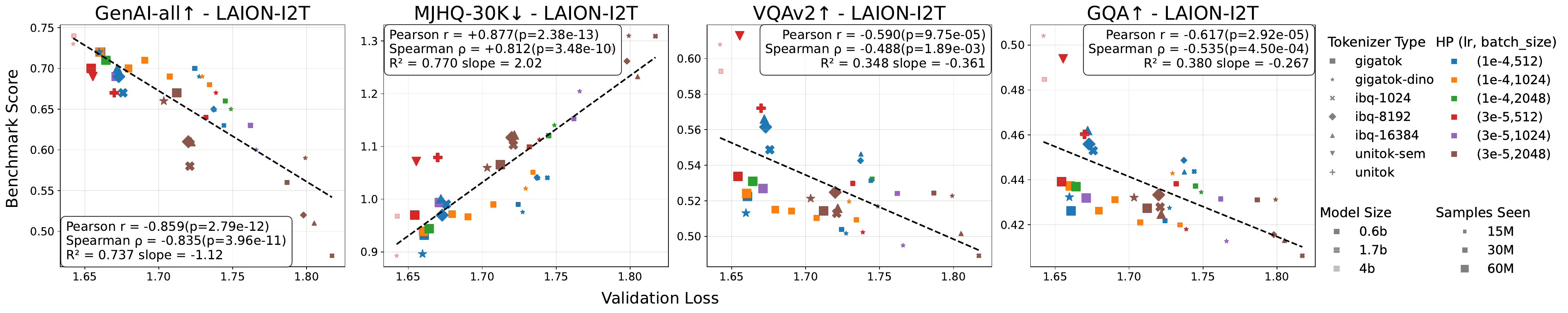}}
    \vskip -0.15in
    \caption{\textbf{I2T loss shows moderate correlation with both generation and understanding performance after SFT across tokenizers.} We use the checkpoints after annealing for loss evaluation.}
    \label{loss-post-joint}
  \end{center}
  \vskip -0.2in
\end{figure}
\textbf{Losses correlate with generation quality after SFT.} The pretraining alignment of Section~\ref{sec_pretraining} carries over to post-SFT performance. When the tokenizer is fixed (GigaTok), both the T2I and I2T validation losses measured before SFT remain strongly correlated with post-SFT benchmark performance (Figure~\ref{loss-genai-post}). Across tokenizers, the I2T loss again remains the more consistent signal, staying correlated with post-SFT generation performance (Figure~\ref{loss-post-joint}).

\textbf{I2T loss is informative about general visual understanding.} Across tokenizers, hyperparameters, and training scales, the I2T loss shows a moderate correlation with post-SFT VQAv2 and GQA performance (Figure~\ref{loss-post-joint}), indicating that it captures part of the capability that transfers to general VQA and can thus serve as a meaningful signal. The moderate correlation may partly reflect differences in both task and training stage: the loss measures caption prediction during pretraining, while the benchmarks measure question answering after SFT. Caption fit is therefore informative about VQA performance, but does not determine it. This imperfect correlation is unlikely to be explained solely by data-shuffling noise: in our reproducibility check of one GigaTok configuration, VQAv2 and GQA scores vary by less than $1\%$ relative across two seeds (Table~\ref{appendix_tab_seed}). The relationship is less consistent for specialized benchmarks such as TextVQA, which additionally require capabilities such as OCR (Appendix~\ref{appendix_loss_textvqa}).

\par
\finding{Pretraining losses are informative about post-SFT performance. I2T loss correlates with both generation and general visual understanding across tokenizers, while the relationship between T2I loss and generation performance is tokenizer-dependent.}

\section{What Unified Training Reveals About Image Tokenizers}
\label{sec_case_study}
Building on our loss-based analysis of unified multimodal training (Section~\ref{sec_loss_interpret}), we now use validation loss as a lens to reveal tokenizer properties that isolated metrics or single-axis evaluations miss: reconstruction fidelity can diverge from multimodal learnability, and the image token space can even affect text modeling (Section~\ref{sec_interference}). We study these effects through three tokenizer design axes as case studies: the discriminator (Section~\ref{sec_discriminator}), semantic supervision (Section~\ref{sec_semantic_loss}), and vocabulary size (Section~\ref{sec_vocabulary_size}).
\subsection{Does better reconstruction imply better unified multimodal learning?}
\label{sec_discriminator}
The premise behind reconstruction-based tokenizer selection is that higher-fidelity reconstruction yields better downstream results. Recent analyses question this premise, showing that tokenizer design trades off reconstruction fidelity, compression, and latent-space learnability, though largely in the visual-generation-only setting~\cite{bfl2025representation, yao2025reconstruction, ramanujan2024worse, xiong2025gigatok}. We revisit this fidelity--learnability--performance relationship in unified AR multimodal training, where we measure fidelity by rFID and take \emph{multimodal learnability} to be how well the resulting image and text tokens are modeled under the shared AR objective, as reflected by the task-specific validation losses of Section~\ref{sec_loss_interpret}.
\setlength{\tabcolsep}{7pt}
\begin{table}[t] \vspace{-8pt} \begin{center} \caption{\textbf{Tokenizer comparison at 0.6B model scale.} (rFID: reconstruction FID on ImageNet-1K; Text/I2T/T2I: last step loss; GenAI and VQAv2 are post-SFT; lr=1e-4 for GigaTok family, lr=3e-5 for UniTok family, batch size=512; $\dagger$: tokenizers trained by us.)} \label{tab4_learnability} \vskip -0.05in { \begin{tabular}{l c @{\hspace{0.9em}} ccc @{\hspace{0.9em}} cc} \toprule \multirow{2}{*}{Tokenizer} & \multicolumn{1}{c}{Fidelity} & \multicolumn{3}{c}{Multimodal Learnability} & \multicolumn{2}{c}{Performance} \\ \cmidrule(lr){2-2} \cmidrule(lr){3-5} \cmidrule(lr){6-7} & rFID$\downarrow$ & Text$\downarrow$ & I2T$\downarrow$ & T2I$\downarrow$ & GenAI$\uparrow$ & VQAv2$\uparrow$ \\ \midrule
      GigaTok-DINO &\textbf{ 0.51} & 2.949 & \textbf{1.660} & 7.437 & \textbf{0.720} & 51.31 \\
      GigaTok & 0.81 & \textbf{2.937} & 1.661 & \textbf{7.416} & \textbf{0.720} & \textbf{52.25} \\
      \hdashline \noalign{\vskip 1pt}
      UniTok$^\dagger$ & \textbf{1.86} & 2.883 & 1.670 & 6.264 & 0.670 & 57.21 \\
      UniTok-sem$^\dagger$ & 2.23 & \textbf{2.873} & \textbf{1.655} & \textbf{5.848} & \textbf{0.690} & \textbf{61.28} \\
      \bottomrule
      \end{tabular}}
    \end{center}
    \vspace{-18pt}           
  \end{table}

\textbf{Better reconstruction does not always lead to better unified multimodal learning.} We compare several tokenizer pairs on reconstruction fidelity, multimodal learnability, and downstream performance in Table~\ref{tab4_learnability}. Within each tokenizer family, better reconstruction does not consistently correspond to better downstream generation or understanding performance. Replacing the standard PatchGAN discriminator with a DINO-based discriminator improves GigaTok's rFID from $0.81$ to $0.51$, yet this reconstruction gain does not translate into a better joint model: none of the validation losses improve significantly, GenAI is unchanged ($0.720$), and VQAv2 drops from $52.25$ to $51.31$. This complements the image-generation-only study of GigaTok~\cite{xiong2025gigatok}, where the DINO-based discriminator is adopted to improve reconstruction. In contrast, semantic supervision in UniTok-sem worsens reconstruction fidelity yet improves all three reported validation losses and both reported downstream metrics (Section~\ref{sec_semantic_loss}). Together, these results show that a reconstruction metric such as rFID alone is insufficient for tokenizer selection: we must also consider how the tokenizer affects multimodal learnability.
\finding{Reconstruction fidelity and multimodal learnability can diverge under unified AR training: lower rFID does not promise stronger downstream generation or understanding.}
\finding{Improving rFID with a DINO-based discriminator does not improve multimodal learnability or downstream performance.}

\subsection{Does the image token space affect text modeling?}
\label{sec_interference}
In unified AR training, text and image tokens are optimized under the same model and objective. This raises a question not addressed by evaluating generation or understanding alone: can the image tokenizer affect text modeling even when the text-only data, text tokenizer, and language backbone are all fixed?

\begin{wrapfigure}[17]{r}{0.49\textwidth}
\vspace{-18pt}
  \centering
    \includegraphics[width=0.49\textwidth]{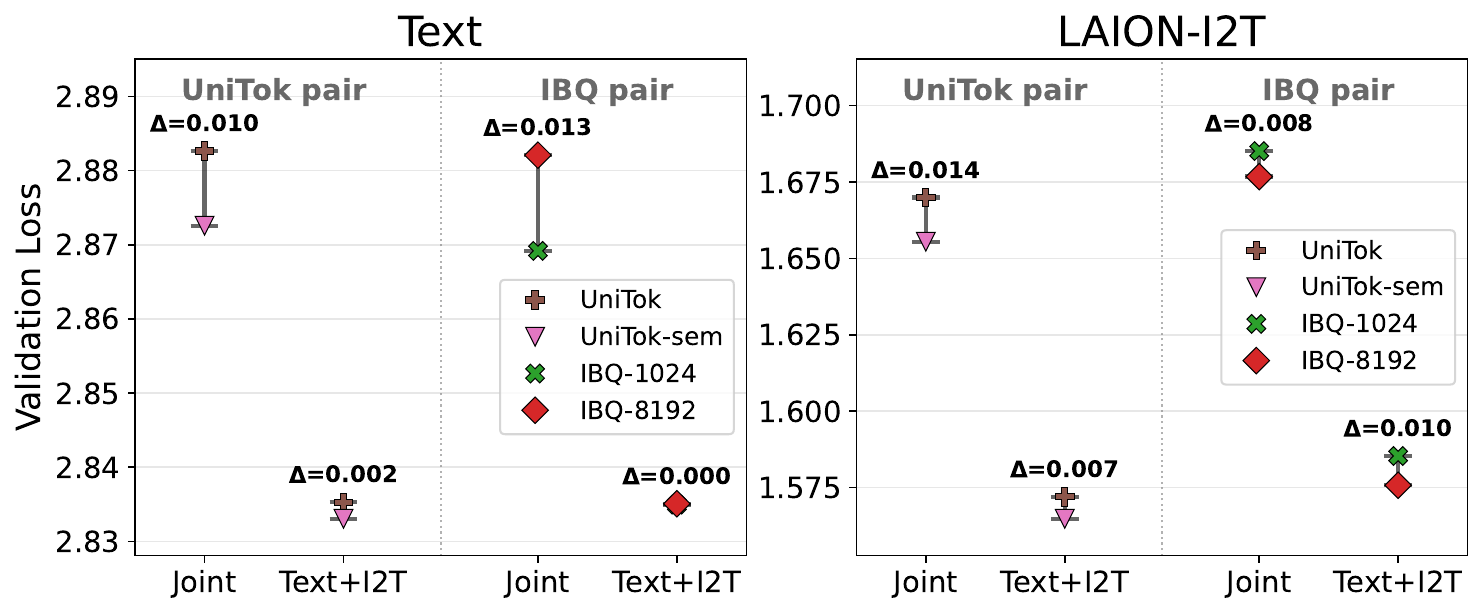}
  \vskip -0.13in
  \caption{\textbf{Changing the image token space can reduce interference between image-token and text-token modeling in joint training.} In the joint setting, UniTok-sem and IBQ-1024 attain lower text loss than UniTok and IBQ-8192, respectively. When the image-generation objective is ablated (Text+I2T training), the text losses within each pair become similar (left), indicating that the interference stems from image-token modeling; the I2T gap, however, persists (right).}
  \label{loss-caption-only}
  \vspace{-0pt}
\end{wrapfigure}
\textbf{Image token space affects text modeling difficulty under joint training.} We observe this effect in two tokenizer comparisons (lr=3e-5, batch size=512) in Figure~\ref{loss-caption-only} (Joint). First, UniTok and UniTok-sem share the same text tokenizer and training recipe, yet UniTok-sem achieves lower text loss under joint text-image training. Second, among IBQ variants, IBQ-1024 obtains lower text loss than IBQ-8192 despite using the same text token space. These differences suggest that changes in the image token space can affect the difficulty of text modeling. In these comparisons, adding semantic supervision to UniTok or reducing the IBQ vocabulary from 8192 to 1024 lowers text loss under joint training.

\textbf{The text-side effect comes from image token prediction rather than captioning (I2T).} We hypothesize that this text-side effect arises from interference between image token modeling and text token modeling. To test this, we remove the image token prediction objective by reformatting the T2I and image-generation samples into I2T order and train on only Text+I2T data with the same hyperparameter settings. Under this ablation, the text-loss gap largely disappears for both the UniTok pair and the IBQ pair (Figure~\ref{loss-caption-only}, Text+I2T), indicating that the text-side difference is explained by the image token prediction objective in full unified training. Interestingly, the I2T gap in both pairs persists, suggesting that tokenizer choice affects how informative image tokens are for captioning through a pathway independent of image token prediction. In Appendix~\ref{appendix_ablation_I2T}, we further verify that the text-side effect is not caused by the I2T objective through a similar ablation on I2T data.
\par 
\finding{Image token space can affect text modeling difficulty under joint AR training. Our ablations attribute this effect to the image token prediction objective rather than to captioning (I2T).}
\subsection{Semantic supervision: How does semantic alignment change multimodal learnability?}
\label{sec_semantic_loss}
Semantic supervision is commonly added in image tokenizer training to encourage visual features to encode higher-level semantics. In Section~\ref{sec_discriminator}, we observed that it improves the multimodal learnability of the resulting visual language, which in turn benefits downstream performance. We now ask where this improvement comes from: does semantic supervision make local image-token sequences easier to predict (``better local visual grammar''), or does it make individual image tokens more informative for captioning (``better visual words'')?

\begin{figure}[h]
  \centering
  \vskip -0.1in
  \begin{minipage}[t]{0.573\textwidth}
    \centering
    \includegraphics[width=\linewidth]{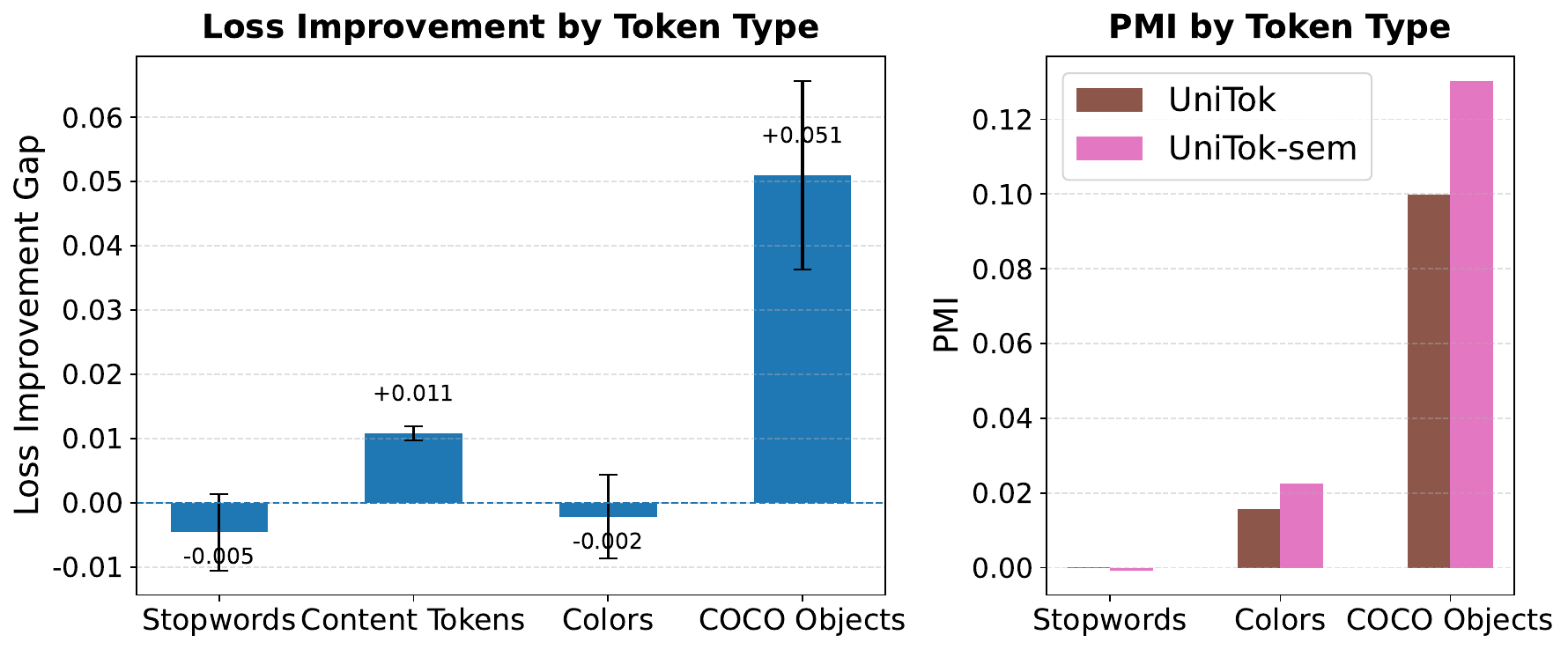}
    \caption{\textbf{Semantic supervision strengthens object-level image-token--word alignment (``better visual words'').} UniTok-sem yields larger I2T loss improvements on COCO object words than on stopwords, colors, or general content tokens (left), and higher PMI between image tokens and object words (right).}
    \label{loss-unitok-analysis}
  \end{minipage}\hfill
  \begin{minipage}[t]{0.407\textwidth}
    \centering
    \includegraphics[width=\linewidth]{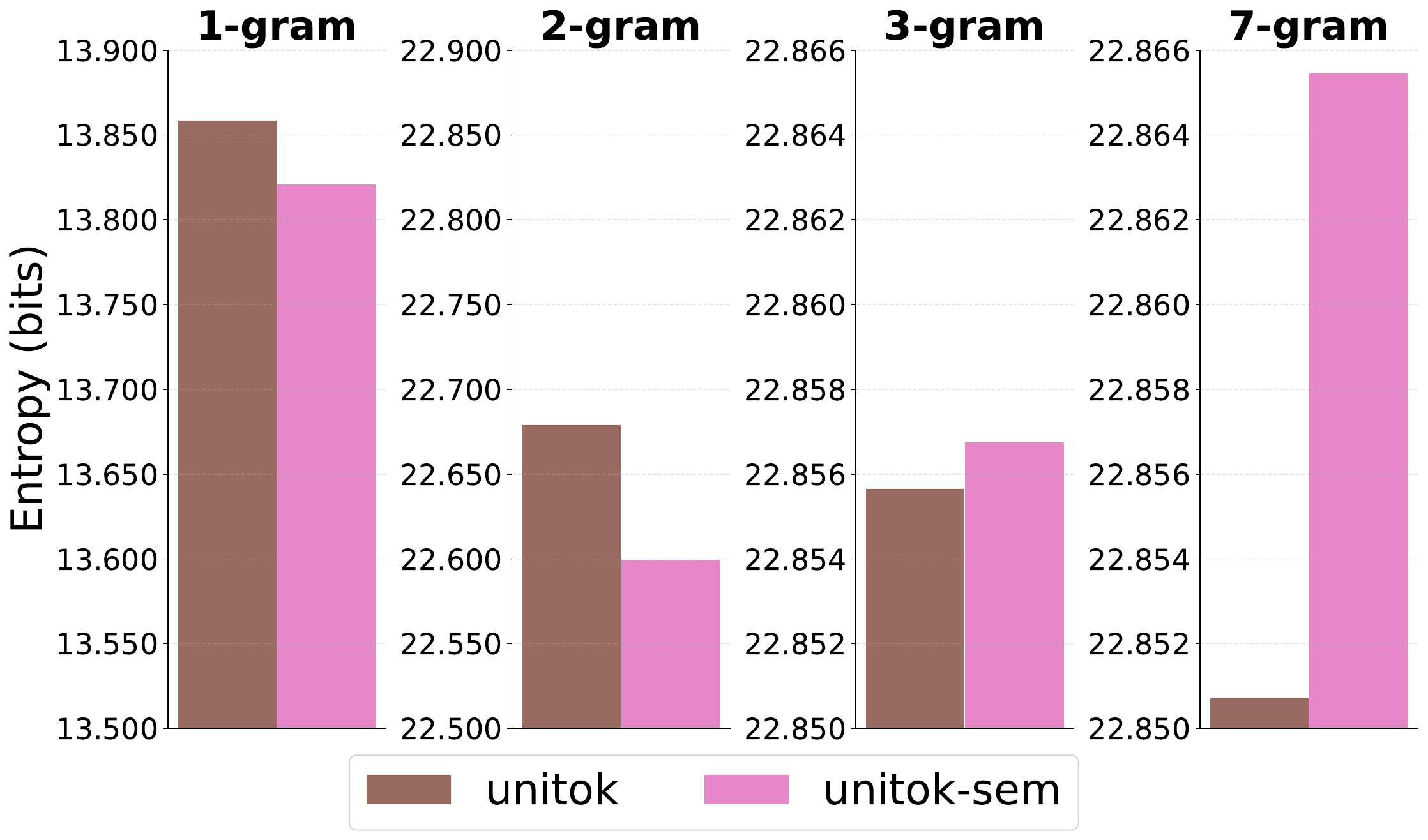}
    \caption{\textbf{Semantic supervision does not consistently reduce empirical image-token $n$-gram entropy.} The empirical entropy of the tokens is calculated on tokenized images of the LAION-Aesthetics validation set.}
    \label{entropy-analysis}
  \end{minipage}
  \vskip -0.1in
\end{figure}
\textbf{The I2T gain is concentrated on visually grounded object words.} In the Text+I2T setting of Section~\ref{sec_interference}, where image token prediction is ablated, we track each tokenizer's I2T loss improvement over training and take the gap ($\Delta_{\text{UniTok-sem}} - \Delta_{\text{UniTok}}$), for which larger values mean greater relative improvement of UniTok-sem. This gain concentrates on visually grounded object words (Figure~\ref{loss-unitok-analysis}, left): COCO object-category tokens improve far more than stopwords, general content tokens (all tokens apart from stopwords), or color tokens (CSS color names), indicating that semantic supervision makes image tokens more informative for predicting object-level words.

\textbf{Semantic supervision strengthens object-level image-token--word association.} As further evidence, the pointwise mutual information (PMI) between image tokens and words, averaged within three word groups on the validation set, is higher for UniTok-sem on COCO object words (Figure~\ref{loss-unitok-analysis}, right), suggesting that semantic supervision associates image tokens more tightly with object-level words.

\textbf{Semantic supervision does not consistently reduce empirical image-token $n$-gram entropy.} An alternative explanation is that semantic supervision makes image-token sequences easier to model. Following \citet{chan2024analyzing}, we compute the empirical 1-, 2-, 3-, and 7-gram entropy of the image tokens produced by UniTok and UniTok-sem on the validation set (Figure~\ref{entropy-analysis}). While UniTok-sem tokens have slightly lower entropy at 1- and 2-grams, their 3- and 7-gram entropy is higher than for UniTok. These measurements do not show a consistent reduction in local $n$-gram entropy, providing no clear evidence for a simpler local visual grammar under this diagnostic.
\par\vspace{-1pt}
\finding{Semantic supervision improves multimodal learnability and strengthens object-level image-token--word associations, without consistently reducing empirical $n$-gram entropy. These observations support a ``better visual words'' interpretation rather than a simpler local visual grammar.}

\subsection{Vocabulary size: Is a larger vocabulary better for unified multimodal learning?}
\label{sec_vocabulary_size}
Vocabulary size is a key compression axis of discrete image tokenizers. Prior work shows that scaling up the vocabulary improves reconstruction fidelity by letting each image token carry more visual information~\cite{shi2025scalable}, but it may also change the modeling difficulty of image tokens. Since reconstruction fidelity and multimodal learnability can diverge, we ask whether a larger vocabulary brings better unified multimodal learning. Within the IBQ family, we vary $B\in\{1024,8192,16384\}$ and compare image-side losses using the vocabulary-normalized image-token loss $\mathcal{L}^{\ast}$ (Section~\ref{sec_pretraining}).

\begin{figure}[h]
  \vskip -0.1in
  \begin{center}
    \centerline{\includegraphics[width=1.0\textwidth]{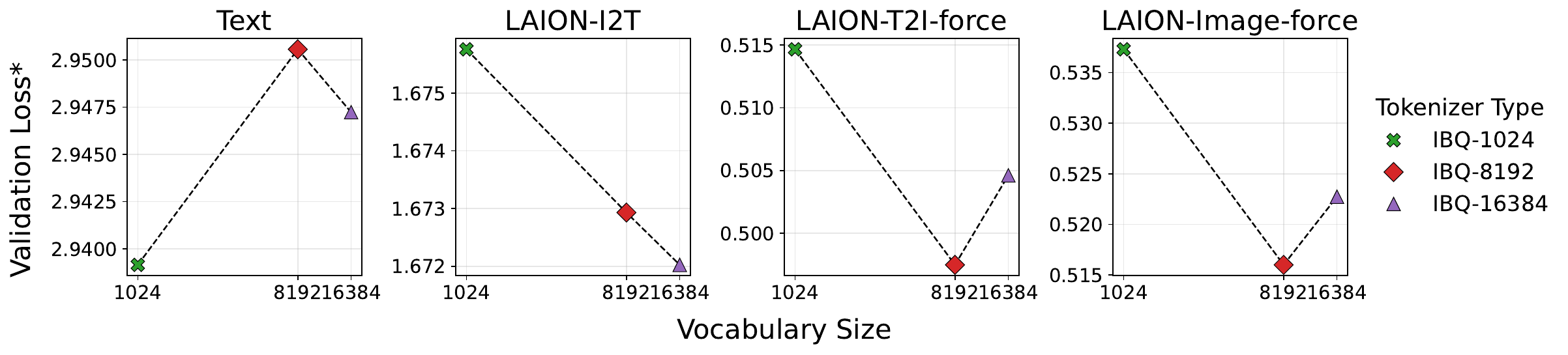}}
    \vskip -0.15in
    \caption{
      \textbf{Vocabulary size changes task-specific validation losses with no simple monotonic trend.} Among the three IBQ variants, IBQ-16384 achieves the best I2T loss, but IBQ-8192 attains the lowest T2I and image losses after normalization (lr=1e-4, batch size=512).
    }
    \label{loss-flops-mixpretrain-ibq}
  \end{center}
  \vskip -0.25in
\end{figure}
\textbf{Vocabulary size affects task-specific losses non-monotonically.} As shown in Figure~\ref{loss-flops-mixpretrain-ibq}, the relationship between vocabulary size $B$ and loss is not simply larger-is-better. IBQ-16384 attains the best I2T loss, yet the intermediate size IBQ-8192 achieves the lowest normalized T2I and image losses among the three variants. Text loss, meanwhile, follows the reverse (also non-monotonic) trend relative to the image-side losses, again indicating that different tasks can prefer different visual token spaces.

\begin{wrapfigure}[10]{r}{0.49\textwidth}
\vspace{-15pt}
  \centering
    \includegraphics[width=0.49\textwidth]{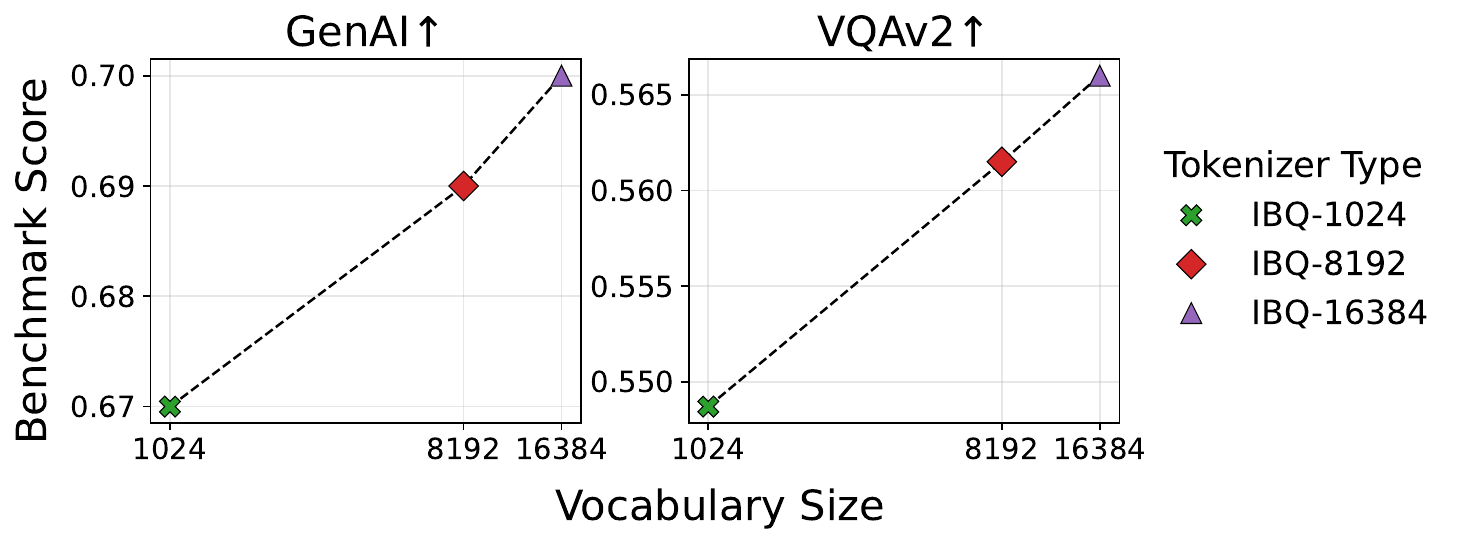}
  \vskip -0.13in
  \caption{\textbf{The largest vocabulary brings the best downstream performance among the three IBQ variants} (lr=1e-4, batch size=512).}
  \label{benchmark-mixpretrain-ibq}
  \vskip -0.22in
\end{wrapfigure}
\textbf{A larger vocabulary can benefit downstream performance, possibly through higher reconstruction fidelity.} Comparing the IBQ variants on generation and understanding benchmarks after SFT (Figure~\ref{benchmark-mixpretrain-ibq}), IBQ-16384 achieves the best performance among the three, despite IBQ-8192 having lower T2I and image losses. As with the rFID--performance relation in Section~\ref{sec_pretraining}, this is plausibly due to IBQ-16384's higher reconstruction fidelity. Beyond $B=16384$, however, it remains unclear whether reconstruction fidelity or multimodal learnability is the dominant factor, which we leave to future work.
\par
\finding{Vocabulary size affects multimodal learnability non-monotonically: an intermediate vocabulary gives the best T2I and image losses, yet a larger vocabulary can still benefit downstream performance, potentially through higher reconstruction fidelity.}
\section{Conclusion}
We study image tokenizers in the context of unified AR multimodal models through a controlled, loss-based testbed. We first examine how validation loss relates to downstream performance and how the task-specific losses should be interpreted. Building on this, we show that reconstruction fidelity can diverge from multimodal learnability and that the image token space can affect text modeling under joint training. We also revisit discriminator choice, semantic supervision, and vocabulary size to examine their effects on unified multimodal learning. Our findings suggest that unified tokenizer design should account for both reconstruction fidelity and joint modeling difficulty across tasks, rather than optimizing any single proxy metric. These observations highlight the importance of studying image tokenizers as visual languages in interaction with text during joint multimodal training.

\begin{ack}
SSD acknowledges the support of  NSF IIS 2143493, NSF IIS 2229881, Sloan Fellowship, and the AI2050 program at Schmidt Sciences.
This material is based on the Chameleon tokenizer supported by the Chameleon Research License, Copyright (c) Meta Platforms, Inc. All Rights Reserved.
\end{ack}

\bibliography{references}
\bibliographystyle{abbrvnat}
\newpage
\appendix
\section{Training Details}\label{appendix_training}
\subsection{Data Proportion}\label{appendix_training_proportion}
We detail the data mixture in Table~\ref{tab6_data_proportion} and \ref{tab7_sft_data_proportion}. The SFT data is similar to Liquid \cite{liquid}, but we substitute the text-to-image part with public data from pretraining and use more DVQA data.

We adopt 1:8 as the text:image-text ratio in continual pretraining. This choice is influenced by 1:2 in Liquid, while they use the Chameleon tokenizer with a sequence length of 1024. We observe that it is important to keep the total number of image tokens seen during training at the same level to achieve comparable performance, so we use 1:8 as the final ratio.  
\setlength{\tabcolsep}{5pt}
\begin{table}[t]
  \caption{Proportion of subsets in continual pretraining data. *We upsample JourneyDB from 4.2M to 6.5M.}
  \label{tab6_data_proportion}
  \begin{center}
    \begin{small}
        \begin{tabular}{lccc}
          \toprule
          Data Source  & \# Data (M) & Filtering & Recaptioning \\
          \midrule
          DataComp-LM    & 6.6 & Random Sampling &  -- \\
          LAION-Aesthetics & 42 & Aesthetic Score (5.5) & InternVL3-1B \cite{zhu2025internvl3} \\
          JourneyDB    & 6.5* & -- & GPT3.5 \\
          BLIP3o-Pretrain-Short-Caption    & 4.8  &  -- & --\\
          \bottomrule
        \end{tabular}
    \end{small}
  \end{center}
  \vskip -0.1in
\end{table}
\begin{table}[t]
  \caption{Proportion of subsets in supervised finetuning data. The LAION and JourneyDB text-to-image data is randomly sampled from the same distribution of corresponding data in continual pretraining. *We keep all 0.2M DVQA data without the filtering in Mini-Gemini. }
  \label{tab7_sft_data_proportion}
  \begin{center}
    \begin{small}
        \begin{tabular}{lc}
          \toprule
          Data Source  & \# Data (M)  \\
          \midrule
          Mini-Gemini (VQA)    & 1.7*\\
          Mini-Gemini (Caption) & 1.2\\
          LMSYS-Chat & 1.0 \\
          LAION-Aesthetics & 0.86\\
          JourneyDB    & 0.08\\
          BLIP3o-60K    &  0.06 \\
          \bottomrule
        \end{tabular}
    \end{small}
  \end{center}
  \vskip -0.1in
\end{table}
\subsection{Prompts}\label{appendix_training_prompt}
For visual generation, we uniformly sample the prompt from the same set of prompts as Liquid during pretraining: ``Generate an image based on this description.'',
        ``Create an image that captures the provided description.'',
        ``Based on the previous text, produce a corresponding image.'',
        ``Please illustrate the above text with a picture.'',
        ``Translate the given description into a image.'',
        ``Construct a visual representation of the above description.'',
        ``Create a image that matches the text.'',
        ``Formulate a visual expression that reflects the narrative just provided.'',
        ``Give a visual depiction based on the above sentences.'',
        ``Create an image using the information mentioned above as guidance.''. During loss calculation, we only use ``Generate an image based on this description.'' Note that for unconditional visual generation, we use the format $\text{⟨unconditional⟩⟨boi⟩\{\textbf{image tokens}\}⟨\textbf{eoi}⟩⟨\textbf{eos}⟩}$ without any prompt.

For captioning, we use ``The caption of this image is:'' as the prompt. 
\subsection{Model Performance}\label{appendix_training_comparison}
We list the model performance on visual generation and understanding benchmarks in Table~\ref{tab3_performance} and provide qualitative image generation results on GenAI-Bench prompts in Figure~\ref{qualitative_8b}. Our 8B model trained on public images with less pretraining data achieves similar performance to Liquid-7B on most benchmarks, which verifies that the framework serves as a reusable and reliable testbed for studying tokenizers' impact on downstream unified training.
\setlength{\tabcolsep}{2pt}
\begin{table}
  \centering
  \caption{Comparison of generation and understanding performance with Liquid-7B \cite{liquid} after SFT. ``Data'' refers to the number of samples seen in continual pretraining.}
   \label{tab3_performance}
   {\small
  \begin{tabular}{lccccccccc}
  \toprule
    Model & Data &GenAI$\uparrow$ & MJHQ-30K$\downarrow$ & WISE$\uparrow$ & VQAv2$\uparrow$ & GQA$\uparrow$ & TextVQA$\uparrow$& POPE$\uparrow$& MME$\uparrow$ \\
    \midrule
    Liquid-7B & 90M &0.72 & 5.47 & 0.41 & 68&56.1 &40.4 &81.1 &1107.2\\
    Ours (8B) & 60M & 0.73 &  10.55& 0.38 & 67.3& 54.44&43.46 & 78.1&1038.12\\
    \bottomrule
  \end{tabular}}
\end{table}

\definecolor{qualok}{HTML}{1B7F5A}
\definecolor{qualbad}{HTML}{B3261E}
\newcommand{\qualwidth}{0.325\textwidth}
\newcommand{\qualpanel}[4]{%
  \begin{minipage}[t]{\qualwidth}%
    \centering
    \includegraphics[width=\linewidth]{figures/qualitative_8b/#3}\\[2pt]
    {\scriptsize\textcolor{#1}{\textbf{#2}}}\\[1pt]
    {\scriptsize #4\par}%
  \end{minipage}%
}
\newcommand{\qualgood}[2]{\qualpanel{qualok}{\checkmark\ Success}{#1}{#2}}
\newcommand{\qualfail}[2]{\qualpanel{qualbad}{$\times$\ Failure}{#1}{#2}}

\begin{figure}[t]
  \centering
  \setlength{\tabcolsep}{2pt}
  \begin{tabular}{@{}ccc@{}}
    \qualgood{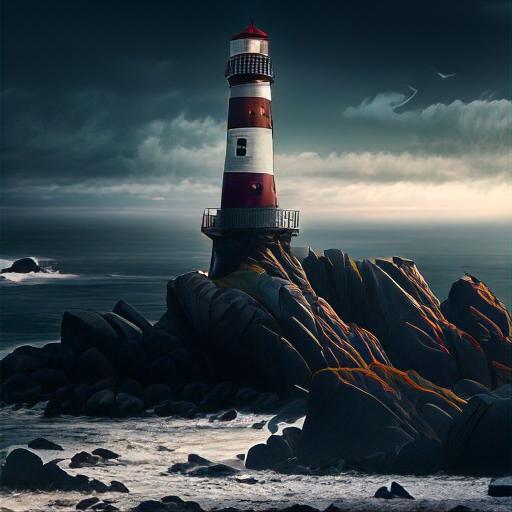}{``A lone lighthouse standing guard on a rocky coastline.''} &
    \qualgood{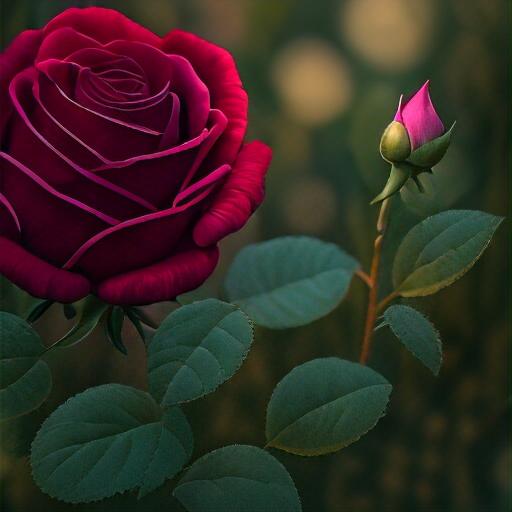}{``A red rose in full bloom next to a pink rosebud in a garden.''} &
    \qualgood{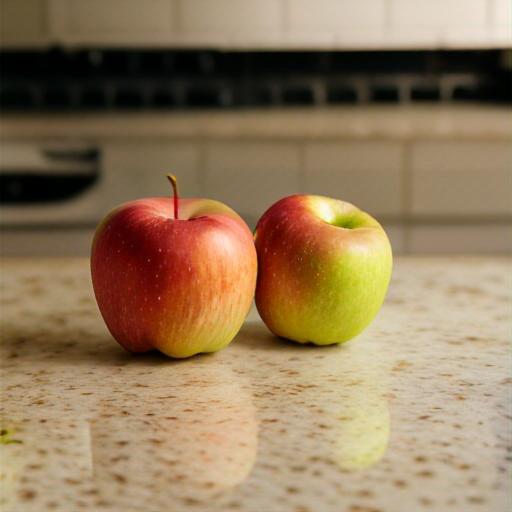}{``Two apples on a kitchen counter.''} \\[6pt]
    \qualgood{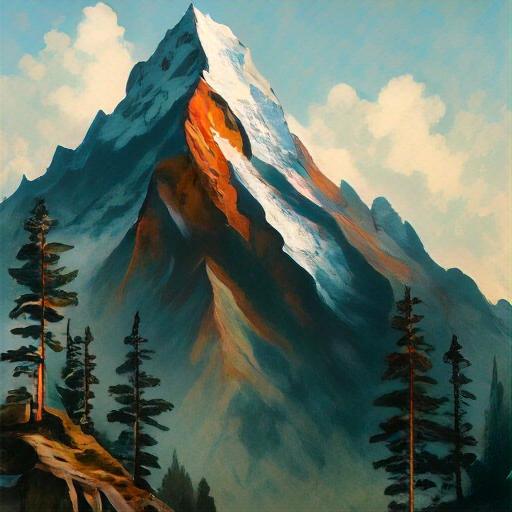}{``A painting where the mountain is depicted as taller than the trees in the foreground.''} &
    \qualfail{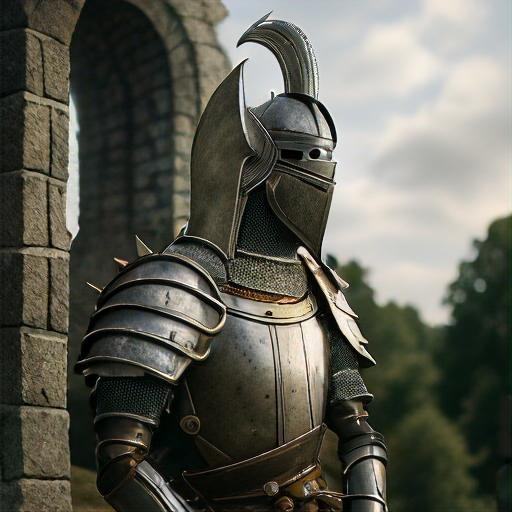}{``A knight with a feather plume helmet by a stone tower.''} &
    \qualfail{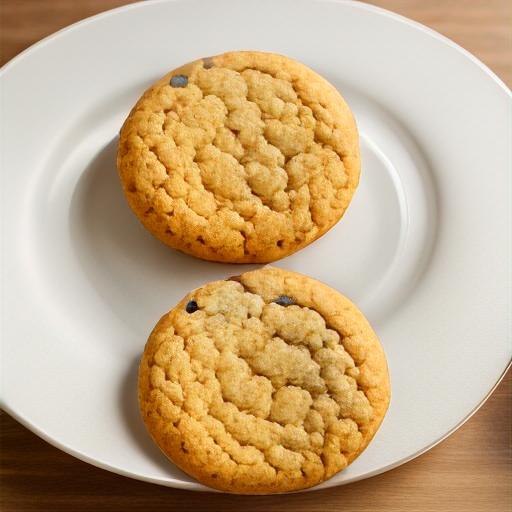}{``Three cookies on a plate.''} \\
  \end{tabular}
  \caption{Qualitative text-to-image generation of our 8B model on GenAI-Bench. All images are generated by the Qwen3-8B model with the Chameleon tokenizer after SFT, i.e., the same checkpoint reported in Table~\ref{tab3_performance}, using the prompt shown below each panel. \textcolor{qualok}{Green} marks cases with successful instruction following and \textcolor{qualbad}{red} marks failure cases. }
  \label{qualitative_8b}
\end{figure}
\subsection{Tokenizer Information}
\label{appendix_tokenizer_information}
IBQ (Index Backpropagation Quantization) \cite{shi2025scalable} modifies standard vector quantization in VQGAN to enable scalable training of large codebooks. Instead of updating only the selected code entries, IBQ backpropagates gradients to all codebook embeddings jointly with the visual encoder, which helps maintain high codebook utilization and a more consistent latent space between encoded features and code vectors. A variant of IBQ is adopted in Emu3.5 \cite{cui2025emu3} with semantic loss and 131072 as the vocabulary size. The authors release IBQ variants with vocabulary sizes of 1024, 8192, 16384, and 262144 without semantic loss. We adopt the first three since 262144 is much larger than the original vocabulary size in the base model, which might cause training difficulty.

GigaTok \cite{xiong2025gigatok} is proposed for autoregressive image generation and studies the effect of tokenizer scale and architectural design on discrete visual representations. It is based on VQGAN with large encoder–decoder networks and incorporates DINO-based semantic loss. It examines design choices such as asymmetric encoder–decoder scaling, different discriminator architectures, and alternative tokenization structures, and reports their impact on reconstruction behavior and downstream image generation. We revisit their conclusion on discriminator architectures using GigaTok-B-L under our framework.

UniTok \cite{ma2025unitok} is a unified discrete image tokenizer designed to support both image generation and visual understanding. It is also based on a VQGAN and introduces multi-codebook quantization, which partitions latent features into multiple subspaces and discretizes each with independent codebooks to increase representational capacity while maintaining stable training. The tokenizer is trained with both reconstruction and semantic loss (contrastive loss with CLIP text encoder) so that the resulting tokens capture fine-grained visual details as well as high-level semantic information. We train a single-codebook version of UniTok with vocabulary size 16384 and latent dimension 64, and ablate the usage of semantic loss, examining its effect on downstream scaling under our framework. Following the original paper, we initialize the model with ViTamin-L/16 \cite{chen2024vitamin}. Due to compute constraints, we train the model on DataComp-medium \cite{gadre2023datacomp} for five epochs and use 4096 as the global batch size. We adopt a smaller learning rate (lr=2e-4 for tokenizer and 5e-6 for discriminator) with entropy loss weight 0.05 and semantic loss weight 0 or 0.05. The rFID, zero-shot accuracy, and linear probing accuracy on ImageNet-1K of the two UniToks are shown in Table \ref{appendix_tab4_unitok_performance}: UniTok-sem is stronger than UniTok in image classification but is weaker in image reconstruction.
\begin{table}
  \centering
  \caption{Comparison of UniTok with and without semantic loss on rFID, top-1 zero-shot accuracy, and top-1 linear probing accuracy on ImageNet-1K. The linear probing training and evaluation follow the same practice as in the GigaTok paper. }
   \label{appendix_tab4_unitok_performance}
   {\small
  \begin{tabular}{lccccccccc}
  \toprule
    Model  &rFID$\downarrow$ & ZS Accuracy$\uparrow$ & LP Accuracy$\uparrow$ \\
    \midrule
    UniTok &1.86 & 5.34 &15.36 \\
    UniTok-sem & 2.23 &  46.79& 59.42\\ 
    \bottomrule
  \end{tabular}}
\end{table}

Following the official code of UniTok, we adopt a zooming-then-cropping strategy for image preprocessing for UniTok and UniTok-sem, which is different from other tokenizers. This strategy is also applied in the pre-tokenization process in our training. Since we do not compare tokenizers from different families, this does not affect our conclusions.
\subsection{Compute Resources}
The experiments were conducted on an internal cluster using 1, 2, 4, or 8 NVIDIA H200 GPUs with 141GB memory. The estimated compute for continual pretraining is listed in the result plots. The full research project also involved preliminary experiments for adjusting the training recipe and hyperparameters.
\section{Supplementary Results on Training Dynamics}
\label{appendix_training_dynamics}
\begin{table}
  \centering
  \caption{Most tokenizers have a linear or concave curve for text but a convex curve for image-related tasks. We list the quadratic coefficient $a$ in quadratic fitting ($y=ax^2+bx+c$) of the 0.6B model runs in Figure~\ref{dataflops-loss}. Positive $a$ indicates the convexity of the curves, while \colorbox{lightgray}{negative $a$} indicates concavity. }
   \label{tab2_convexity}
   {\small
  \begin{tabular}{lcccc}
  \toprule
    Tokenizer & Text & Caption & T2I & Image \\
    \midrule
    GigaTok & 5.46e-04&1.96e-02  & 6.92e-03 &  5.15e-03\\
    GigaTok-DINO & 1.21e-04& \colorbox{lightgray}{-7.73e-03} & 3.66e-03 & 2.54e-03\\
    IBQ-1024 & \colorbox{lightgray}{-1.63e-03} & 1.11e-02 & 2.16e-02 & 1.57e-02\\
    IBQ-8192 & \colorbox{lightgray}{-2.24e-03} & 6.94e-03 & 1.21e-02 & 1.08e-02\\
    IBQ-16384 & \colorbox{lightgray}{-2.17e-03} &8.35e-03& 1.15e-02 & 9.01e-03\\
    UniTok & \colorbox{lightgray}{-1.59e-03}&1.01e-02&1.26e-02&1.15e-02\\
    UniTok-sem & \colorbox{lightgray}{-2.01e-03}&2.19e-02&1.13e-02&1.22e-02\\
    \bottomrule
  \end{tabular}}
\end{table}
\begin{figure}
  \begin{center}
    \centerline{\includegraphics[width=1.0\textwidth]{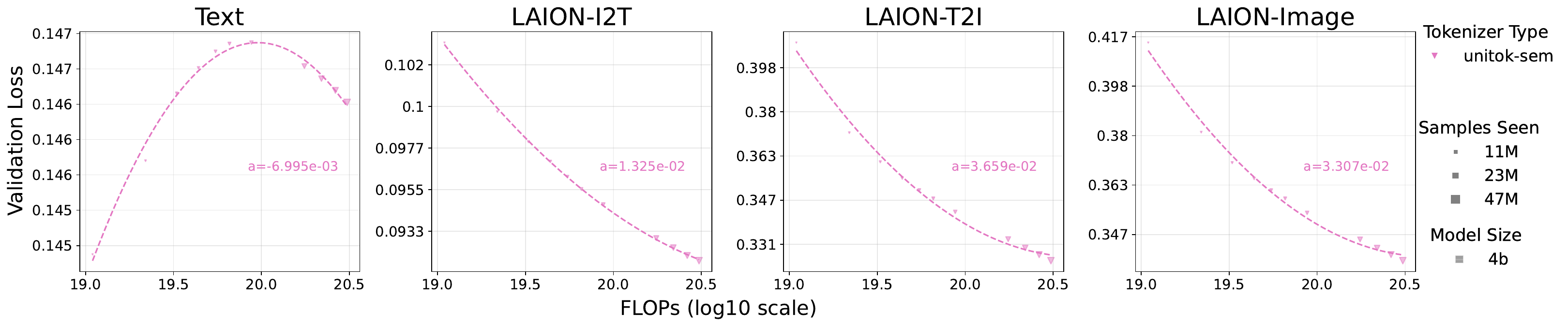}}
    \caption{
      Quadratic fitting of UniTok-sem 4B training, including data points from earlier stage of continual training.
    }
    \label{quad-fitting}
  \end{center}
\end{figure}
\begin{figure}
  \begin{center}
    \centerline{\includegraphics[scale=0.4]{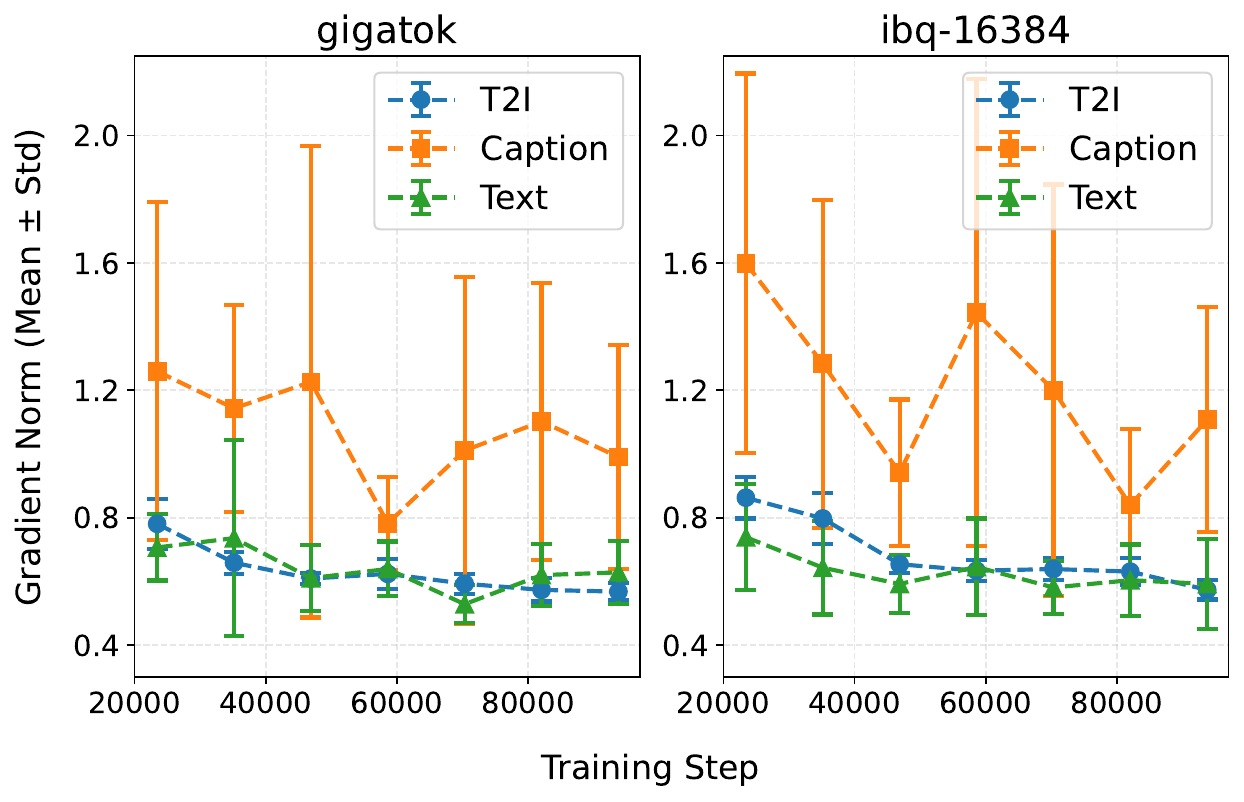}}
    \caption{
      Gradient norms over all parameters by modality across training steps of 0.6B model with GigaTok and IBQ-16384 (lr=3e-5, bs=512). For both runs, the T2I gradient slowly saturates at the end. 
    }
    \label{grad_norm}
  \end{center}
  \vskip -0.3in
\end{figure}
We try quadratic fitting for the loss-data relationship in Figure~\ref{dataflops-loss} for all tokenizers and record the quadratic coefficients in Table~\ref{tab2_convexity}: Most data scaling curves show convexity for LAION-I2T, LAION-T2I, and LAION-Image loss, but a slight concavity for text loss. Considering that the data scaling dots are checkpoints from the same runs, this implies different training dynamics across tasks. In Figure~\ref{quad-fitting}, we present the loss curves of a 4B model training including earlier checkpoints, and find the curve shapes aligned with the finding on quadratic coefficients. We hypothesize that the image-related modeling might dominate optimization in the early phase but slowly saturate later, giving rise to the text loss decaying rate. 

We look into the gradient norm during a single run and check how gradient norm per modality changes over time in continual pretraining with GigaTok and IBQ-16384. In Figure~\ref{grad_norm}, we find that the T2I-related gradient is larger than the text gradient at the beginning, but gradually decays throughout the training. The text gradient decays at a slower rate and reaches a similar level to the T2I gradient at the end. This supports the hypothesis about the convexity and concavity we observe in data scaling curves.
  
\section{Supplementary Loss Results}
\subsection{Loss over Different Image Sources}
\label{appendix_image_loss}
\begin{figure}[ht]
    \vskip 0.0in
  \begin{center}
    \centerline{\includegraphics[width=1.0\textwidth]{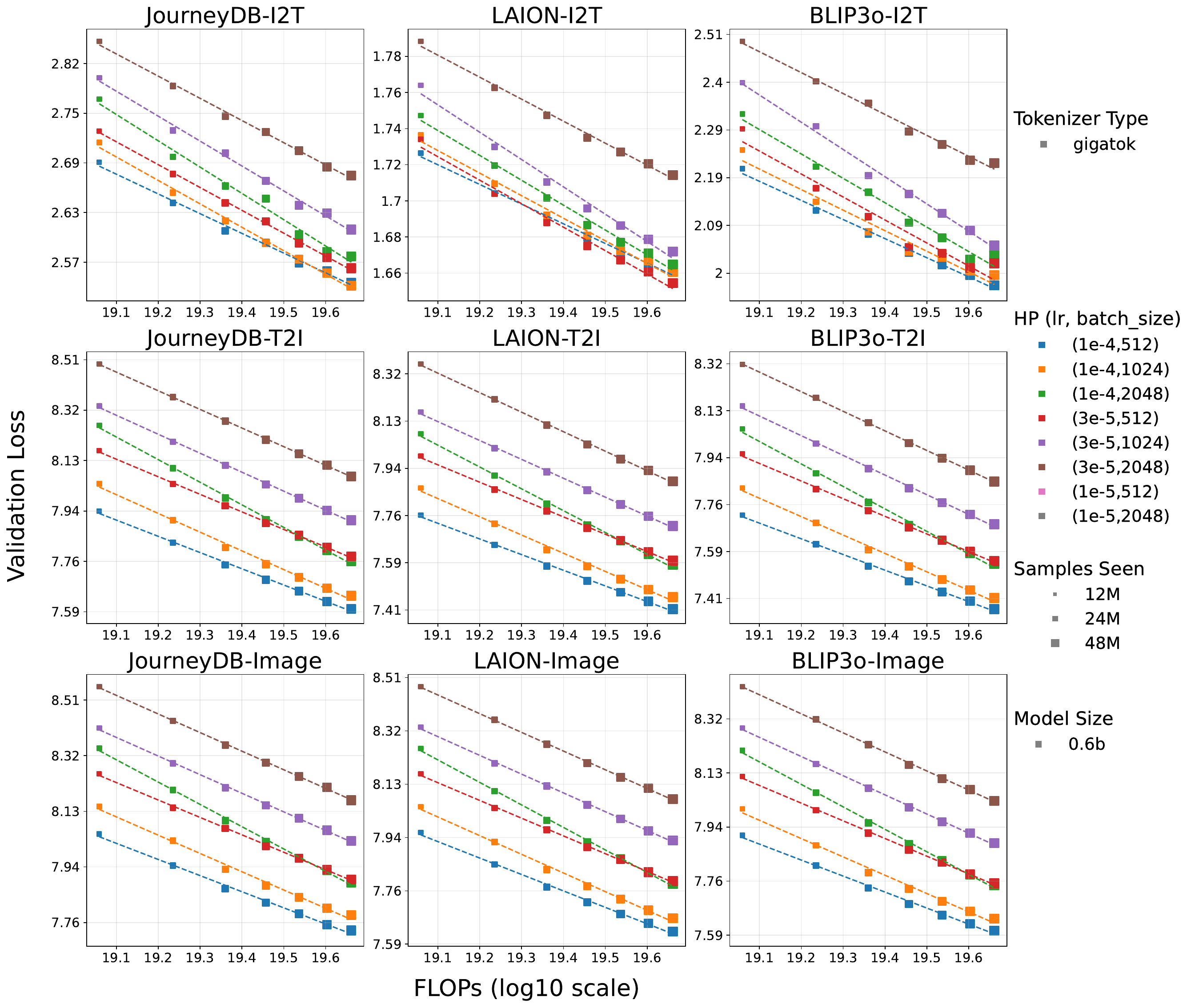}}
    \caption{Loss scales with data similarly for image-text data from different sources.
    }
    \label{flops-loss-hp-sources}
  \end{center}
\end{figure}

We observe very similar T2I loss and image loss trends across three data sources in Figure~\ref{flops-loss-hp-sources}. The I2T loss trends are also similar between JourneyDB and BLIP3o-Short-Caption, but are slightly different from LAION-I2T, where (lr=3e-5, bs=512) surpasses other settings. We hypothesize that this is due to their different caption styles: recaptioned LAION-Aesthetics has longer and more detailed captions than the other two. Therefore, it would rely more on the models' language ability. 
\subsection{Loss Scaling Results for Different Hyperparameters}
\label{appendix_hyperparameters}
\begin{figure}[ht]
  \begin{center}
    \centerline{\includegraphics[width=1.0\textwidth]{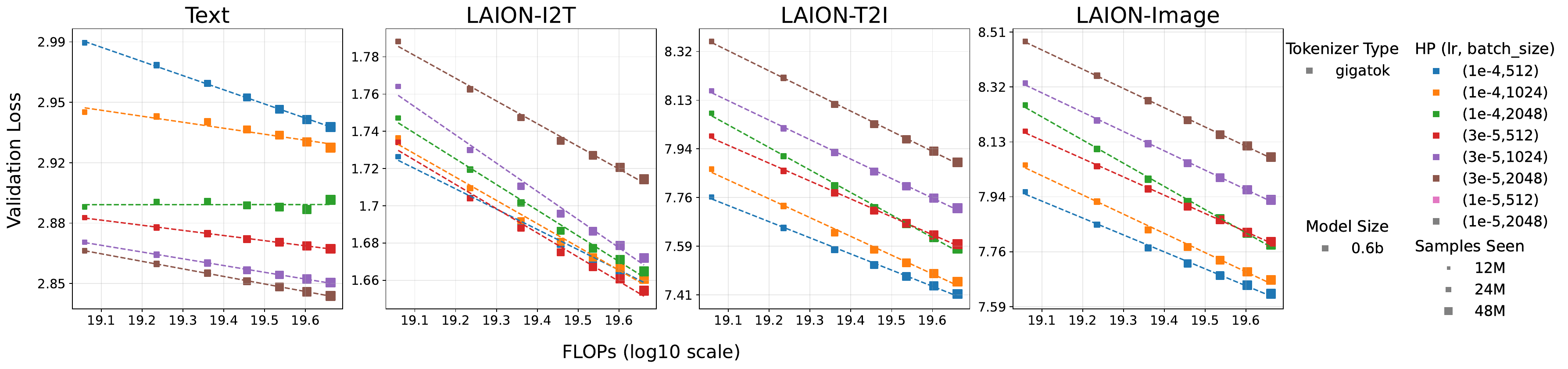}}
    \caption{Loss scales with data for the 0.6B model with GigaTok differently under six hyperparameter settings. Text loss prefers a small learning rate and a large batch size, while image-related tasks favor the opposite settings. 
    }
    \label{flops-loss-hp}
  \end{center}
  \vskip -0.35in
\end{figure}
To study the effect of hyperparameters on model fitting, we experiment with a 2\texttimes 3 grid of (lr, bs) using GigaTok. In Figure~\ref{flops-loss-hp}, we observe that no hyperparameter setting dominates all tasks: Aggressive hyperparameter settings (high lr, small bs, represented by (1e-4,512)) are favored on LAION-I2T, LAION-T2I, and LAION-Image, but lags on text loss. Mild hyperparameters represented by (3e-5, 2048) maintain the property of the base model with better text fitting. Such conflict in hyperparameter preference echoes observations on the trade-off between maintaining the pure-text ability of the base model and accommodating it for image-related tasks \cite{liquid}. 
\subsection{Annealed Loss Results}
\label{appendix_annealed_loss}
\begin{figure}[ht]
    \vskip 0.0in 
  \begin{center}
    \centerline{\includegraphics[width=1.0\textwidth]{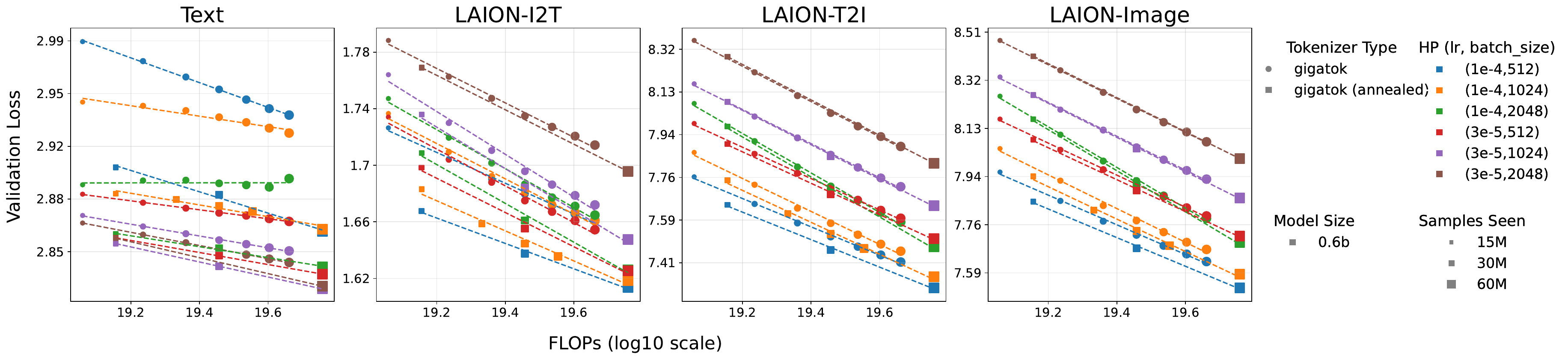}}
    \caption{
      Loss scales with data for annealed and non-annealed checkpoints. 
    }
    \label{flops-loss-hp-annealed}
  \end{center}
\end{figure}
We plot the loss-FLOPs results after annealing along with the non-annealed results in Figure~\ref{flops-loss-hp-annealed}. On all tasks, the relative ranking among hyperparameters is barely changed after annealing compared with results before annealing in Figure~\ref{flops-loss-hp}. One notable exception is that lr=1e-4 has a greater loss decay during annealing: In LAION-I2T, (lr=3e-5, bs=512) reaches the lowest loss at the end of the constant learning rate stage, but is outperformed by (lr=1e-4, bs=512) after annealing. We observe that annealing mainly reduces the text loss but does not affect the image-related losses significantly. One hypothesis is that the loss on image tokens might lead to noisier or smaller gradients during the annealing stage than the loss on the text tokens. 

\subsection{Loss-VQA Performance on TextVQA}
\label{appendix_loss_textvqa}
\begin{wrapfigure}[12]{r}{0.49\textwidth}
\vspace{-20pt}
  \vskip 0.0in
  \begin{center}
    \centerline{\includegraphics[width=0.49\textwidth]{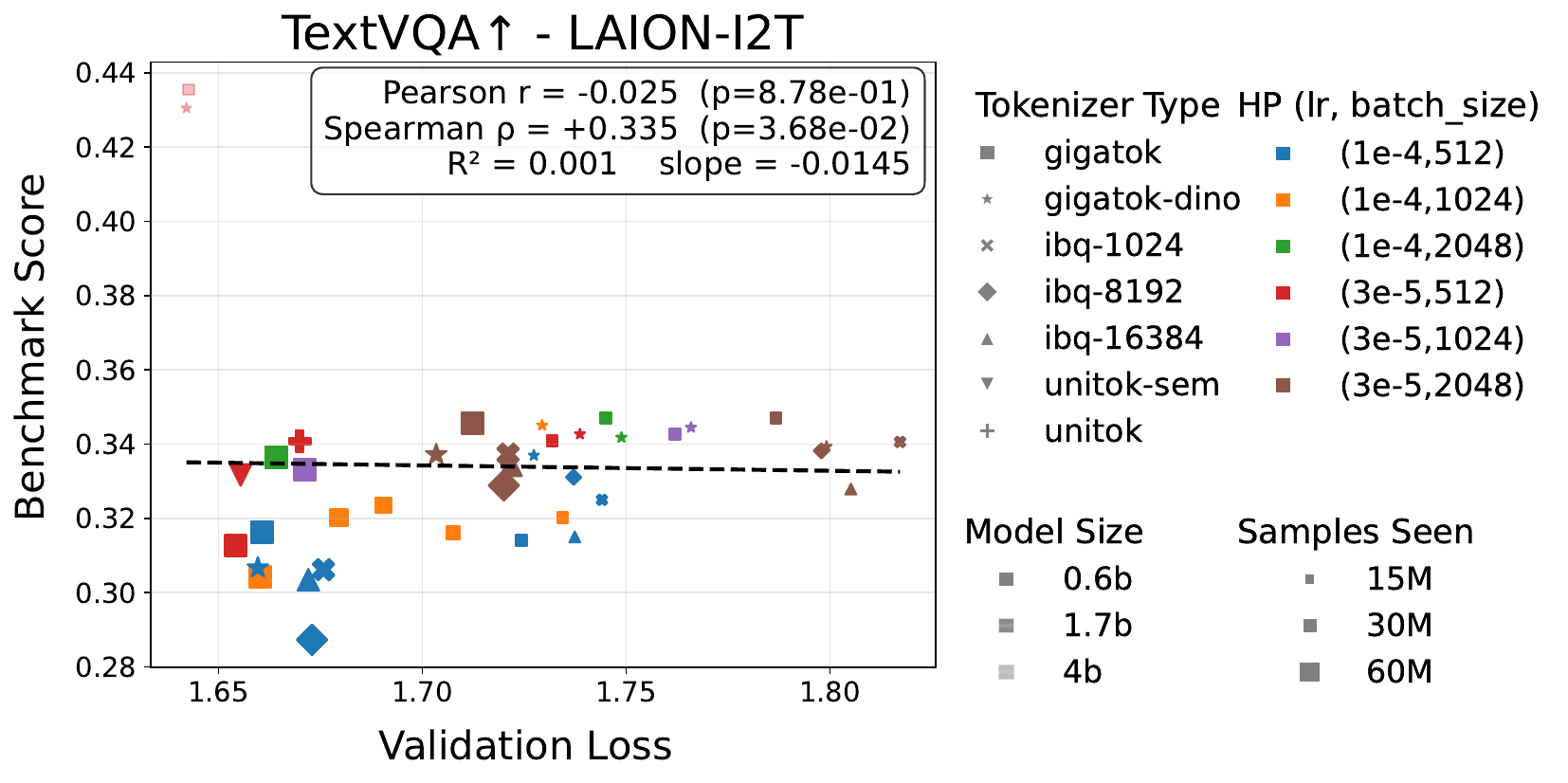}}
    \caption{I2T loss has a noisy and inconsistent correlation with TextVQA performance.}
    \label{loss-textvqa}
  \end{center}
  \vskip -0.3in
\end{wrapfigure}
\par
We study the relationship between pretraining loss and a more specific VQA task: TextVQA. In Figure~\ref{loss-textvqa}, I2T loss is negatively correlated for 0.6B models, but shows the opposite trend when we scale up the model size to 4B. As a side finding, we observe that optimal hyperparameters vary for different benchmarks: Aggressive hyperparameters win on VQAv2 and GQA, but mild hyperparameters lead to better TextVQA performance despite higher I2T loss. These results suggest that post-SFT understanding performance sometimes depends on additional capabilities beyond pretraining caption fit, such as OCR. On the other hand, treating the average score on several VQA benchmarks as the visual understanding performance might omit their distinct properties and could be highly sensitive to benchmark choices. 
\subsection{Details on Loss Normalization}
\label{appendix_loss_normalization}
For LAION-T2I-force and LAION-Image-force, we first set the output logits for non-image tokens to -inf between ⟨boi⟩ and ⟨eoi⟩. This ensures that the model can only choose from the image tokens. Then, we normalize the loss (log perplexity) by log vocabulary size. Empirically, we find that the effect of the first step (forcing image output) is negligible for the loss values, indicating that the model has learned to output image tokens between ⟨boi⟩ and ⟨eoi⟩ on the validation set even at the first checkpoint plotted in the results (trained with 12M data).

Note that we do not apply the same normalization to the text loss and I2T loss even though the total vocabulary sizes are not identical for the three IBQ variants. The reason is that we also find the first step (forcing text output) has a negligible effect on the loss values, which shows that the model only chooses from text tokens during text and I2T evaluation on the validation set. Hence, we do not normalize the losses by the total vocabulary size.

\textbf{Comparison with empirical code entropy.} Normalizing by $\log_2 B$ assumes the maximal entropy of the image vocabulary, whereas real codebooks are used unevenly. To test whether $\log_2 B$ is an adequate stand-in for the actual code distribution, we compare it with the empirical unigram entropy $H_1$ measured on the validation image tokens of all seven tokenizers (Table~\ref{appendix_tab_entropy_norm}). $H_1$ lies within $1.3\%$ of $\log_2 B$ for every tokenizer. Uneven code usage does occur: GigaTok's rank-frequency distribution is visibly skewed, with head codes used roughly $40$ times as often as tail codes. This skew is nevertheless confined to a small fraction of code ranks and carries little probability mass, so GigaTok's entropy stays close to $\log_2 B$. The larger deficits of IBQ-16384 and UniTok-sem instead reflect lower utilization across a broader portion of the codebook.
\begin{table}[t]
  \centering
  \setlength{\tabcolsep}{6pt}
  \caption{Maximal entropy $\log_2 B$ versus empirical unigram entropy $H_1$ of the image tokens on the validation set. $H_1$ is within $1.3\%$ of $\log_2 B$ for all seven tokenizers.}
  \label{appendix_tab_entropy_norm}
  {\small
  \begin{tabular}{lccc}
    \toprule
    Tokenizer & $\log_2 B$ & $H_1$ & $H_1/\log_2 B$ \\
    \midrule
    GigaTok      & 14.000 & 13.953 & 0.997 \\
    GigaTok-DINO & 14.000 & 13.949 & 0.996 \\
    IBQ-1024     & 10.000 &  9.977 & 0.998 \\
    IBQ-8192     & 13.000 & 12.947 & 0.996 \\
    IBQ-16384    & 14.000 & 13.841 & 0.989 \\
    UniTok       & 14.000 & 13.859 & 0.990 \\
    UniTok-sem   & 14.000 & 13.821 & 0.987 \\
    \bottomrule
  \end{tabular}}
\end{table}

\textbf{Substituting $H_1$ for $\log_2 B$.} We then relate the normalized T2I loss to GenAI-all at the 0.6B scale for the three IBQ variants under both normalizations (Table~\ref{appendix_tab_norm_compare}). The two choices give nearly identical results: both recover a strong loss--performance relation, whereas the unnormalized loss explains essentially none of the benchmark variance and tokenizer identity accounts for most of it.
\begin{table}[t]
  \centering
  \setlength{\tabcolsep}{6pt}
  \caption{Relating normalized T2I loss to GenAI-all across the three IBQ variants (0.6B models). Normalizing by the empirical entropy $H_1$ is nearly equivalent to normalizing by $\log_2 B$, while the unnormalized loss leaves the variance to be explained by tokenizer identity.}
  \label{appendix_tab_norm_compare}
  {\small
  \begin{tabular}{lccc}
    \toprule
    Normalization & Pearson $r$ & $R^2$ from loss & Additional $R^2$ from tokenizer identity \\
    \midrule
    None                    & $+0.016$ & 0.000 & 0.972 \\
    Divide by $\log_2 B$    & $-0.957$ & 0.915 & 0.048 \\
    Divide by $H_1$         & $-0.953$ & 0.909 & 0.055 \\
    \bottomrule
  \end{tabular}}
\end{table}

\textbf{Scope of the normalization.} This normalization corrects the loss scale associated with vocabulary size; it does not calibrate losses across tokenizer architectures at a fixed $B$, which is consistent with the cross-tokenizer clustering in Figure~\ref{loss-genai}. Our experiments also do not cover tokenizers with severe or complete codebook collapse; in such a regime the nominal vocabulary size would no longer represent the effective prediction space, and $\log_2 B$ normalization may not be applicable.
\subsection{Ablation on I2T Objective}
\label{appendix_ablation_I2T}
\begin{wrapfigure}[16]{r}{0.49\textwidth}
\vspace{-18pt}
  \centering
    \includegraphics[width=0.49\textwidth]{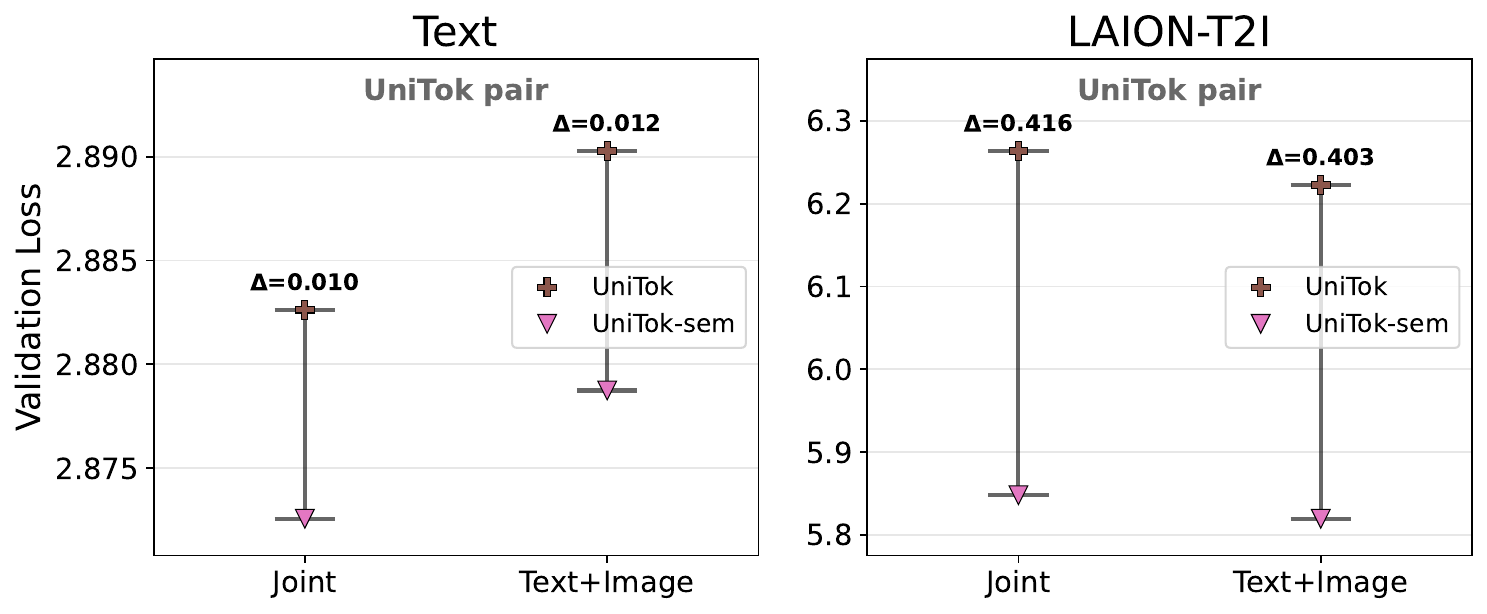}
  \vskip -0.13in
  \caption{In the joint modeling setting, UniTok-sem has lower text loss than UniTok. In Text+T2I training ablating the I2T task, the gap is not reduced, suggesting that the interference does not come from the I2T task. As a side finding, the I2T task complements text modeling in joint training for both tokenizers.}
  \label{loss-image-only}
  \vspace{-0pt}
\end{wrapfigure}
To verify the effect of the I2T objective in joint training, we ablate it by reformatting the I2T samples into T2I order (with drop rate $10\%$ which enables CFG) and train with only Text+T2I data with the same hyperparameter settings. This experiment setting resembles the ablation in Section~\ref{sec_interference} for studying the effect of the T2I objective. 

In Figure~\ref{loss-image-only}, we observe that the gap between tokenizers on text and T2I is not reduced by ablating I2T data. Notably, the text loss increases for both UniTok and UniTok-sem when the I2T objective is ablated. This suggests that the I2T objective complements, rather than competes with, text modeling during joint training. As a comparison, the T2I loss slightly decreases when there is no I2T task. 
\section{Supplementary Benchmark Information and Results}
\label{appendix_generation_benchmarks}
\begin{figure}[ht]
  \vskip 0.2in
  \begin{center}
    \centerline{\includegraphics[width=1.0\textwidth]{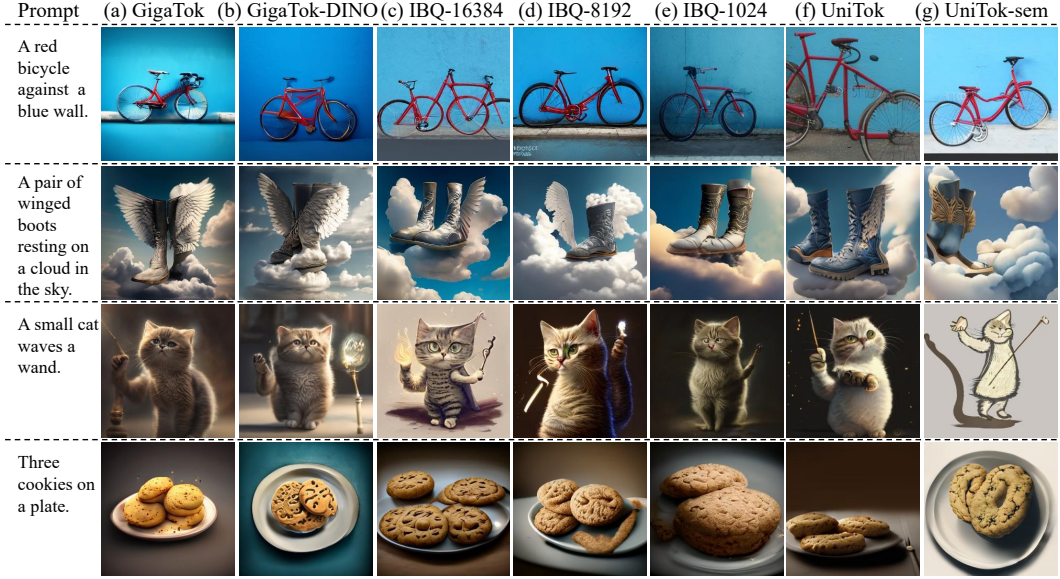}}
    \caption{
      Qualitative comparison of generation quality on GenAI-Bench.
    }
    \label{generation}
  \end{center}
\end{figure}
\textbf{(1) Text-to-image generation.} Following Liquid \cite{liquid}, we report (1) VQAScore on GenAI-Bench \cite{li2024genai}, calculated by prompting CLIP-FlanT5-XXL with whether the image aligns with the text and recording the probability of answering ``Yes'' (hence ranging from 0 to 1), and (2) gFID on MJHQ-30K \cite{li2024playground}.%
During inference, we use unfiltered softmax sampling and CFG scale=7.0. We adopt a unified template ``\{text\} Generate an image based on this description.⟨boi⟩'' for evaluation. Note that we force the model output to be an image token by setting the probability of text tokens to zero during sampling. Without this operation, we find that an under-optimized unified model could output mixed text tokens and image tokens, which even holds for larger public models like Liquid-7B. Some prompts and generated results are in Figure~\ref{generation}. The images are generated by the Qwen3-0.6B model trained with lr=1e-4, bs=512 on 60M data. The qualitative results align with the reported benchmark scores, as GigaTok and GigaTok-DINO produce images with the best quality and alignment, followed by UniTok and UniTok-sem with worse quality but good alignment.

\textbf{(2) Visual understanding.} We use VQAv2 \cite{goyal2017making} and GQA \cite{hudson2019gqa} as two general VQA benchmarks. TextVQA \cite{singh2019towards} is included in Appendix~\ref{appendix_loss_textvqa} as a specific VQA benchmark. We also test the models on POPE \cite{li2023evaluating} and MME \cite{yin2024survey} following Liquid, but find large variance among runs with different shuffling seeds (see Table~\ref{appendix_tab_seed} and analysis below). Therefore, we exclude them from our analysis.

\textbf{Benchmark sensitivity to random seeds.} All central experiments in this work use a single data-shuffling seed (42), which controls the ordering of training data while the language-model initialization is unchanged. To quantify how much of the benchmark spread this ordering accounts for, we additionally ran SFT on one GigaTok configuration (0.6B model, lr=1e-4, bs=1024) with data-shuffling seed 37 and compared the two runs (Table~\ref{appendix_tab_seed}). VQAv2, GQA, GenAI-all, and MJHQ-30K agree to within $1.9\%$ relative difference, whereas POPE changes by $7.07\%$ and MME-P by $4.98\%$. This quantitatively motivates excluding POPE and MME-P from the loss-correlation analysis: their data-order sensitivity is substantially larger than that of the benchmarks we do analyze. We note that this is a two-seed comparison for a single GigaTok configuration rather than a multi-seed replication of every tokenizer comparison due to high computational cost.
\begin{table}[t]
  \centering
  \setlength{\tabcolsep}{6pt}
  \caption{Effect of the data-shuffling seed on post-SFT benchmark scores for one GigaTok configuration (0.6B, lr=1e-4, bs=1024). Relative difference is the absolute difference divided by the mean of the two runs. POPE and MME-P are markedly more sensitive to random seeds than the other benchmarks.}
  \label{appendix_tab_seed}
  {\small
  \begin{tabular}{lccrr}
    \toprule
    Benchmark & Seed 42 & Seed 37 & Abs.\ diff. & Rel.\ diff. \\
    \midrule
    VQAv2$\uparrow$                & 51.43   & 51.47   & 0.04    & 0.08\% \\
    GQA$\uparrow$                  & 43.11   & 42.71   & 0.40    & 0.93\% \\
    POPE$\uparrow$                 & 61.40   & 65.90   & 4.50    & 7.07\% \\
    MME-P$\uparrow$                & 825.56  & 785.48  & 40.08   & 4.98\% \\
    GenAI-all$\uparrow$            & 0.710   & 0.700   & 0.010   & 1.42\% \\
    MJHQ-30K gFID$\downarrow$      & 9.2558  & 9.0812  & 0.1746  & 1.90\% \\
    \bottomrule
  \end{tabular}}
\end{table}
\section{Limitations}
First, our study is limited to seven discrete image tokenizers with fixed token length and single-codebook designs in consideration of a controlled experiment setting. For example, on the sequence length ($K$) axis, we observe that the mixed-modal training is sensitive to the number of image tokens seen. Since it is mathematically impossible to simultaneously hold both the number of image tokens seen and the number of images seen constant when $K$ varies, such a study would introduce confounding variables. We leave the study of tokenizers with sub-codebooks \cite{lu2025atoken}, larger input resolutions \cite{team2024chameleon}, and different sequence lengths \cite{yu2024image} to future research.

Second, we use pretraining validation loss as a lens for studying joint modeling behavior, not as a replacement for downstream evaluation after SFT or for tokenizer selection. Although I2T loss provides a useful signal, it is less predictive for specialized tasks such as TextVQA. 

Third, due to the high computational cost, we do not conduct a full scaling law study on unified multimodal AR training that starts from scratch with more model scales or derive compute-efficient frontiers. Future work could derive and compare image tokenizers' effects on the compute-efficient frontier of multimodal AR models.

Fourth, analyses on the validation set, including PMI and $n$-gram entropy, are diagnostic rather than mechanistic, and high-order empirical entropy can suffer from finite-sample bias.

Fifth, our empirical findings are established for Qwen3-based pure-AR unified models. Holding the backbone family fixed is itself a control in our study, since it lets us vary the visual token space while keeping the downstream architecture and training recipe consistent. We therefore do not claim that every loss relationship or tokenizer ranking is unchanged under a different backbone family; validating them would require matched retraining and evaluation across tokenizers under that backbone. The analysis framework itself can be applied to another language-model family, and we view such cross-family validation as an important direction for future work.

Finally, our downstream evaluation covers a limited set of generation and VQA benchmarks; broader task, safety, and human-preference evaluations are needed to fully evaluate the models. In particular, our understanding evaluation covers general VQA (VQAv2 and GQA) and the OCR-oriented TextVQA, but includes no dedicated region-level grounding or fine-grained spatial-reasoning benchmark. We therefore do not claim that our findings extend to these specialized visual capabilities, and evaluating how the image token space affects them remains future work.

\section{Third-Party Attribution Notices}
Portions of this work use third-party software released under the MIT License:
we train UniTok and UniTok-sem variants using the official UniTok code
(\url{https://github.com/FoundationVision/UniTok}), and we use the GigaTok and
GigaTok-DINO tokenizers (\url{https://github.com/SilentView/GigaTok}). The
copyright and permission notices below are reproduced as required by that
license.

\begin{quote}\small
Copyright (c) 2024 FoundationVision

Copyright (c) 2025 GigaTok: Scaling Visual Tokenizers to 3 Billion Parameters
for Autoregressive Image Generation Authors

Permission is hereby granted, free of charge, to any person obtaining a copy of
this software and associated documentation files (the ``Software''), to deal in
the Software without restriction, including without limitation the rights to
use, copy, modify, merge, publish, distribute, sublicense, and/or sell copies of
the Software, and to permit persons to whom the Software is furnished to do so,
subject to the following conditions:

The above copyright notice and this permission notice shall be included in all
copies or substantial portions of the Software.

THE SOFTWARE IS PROVIDED ``AS IS'', WITHOUT WARRANTY OF ANY KIND, EXPRESS OR
IMPLIED, INCLUDING BUT NOT LIMITED TO THE WARRANTIES OF MERCHANTABILITY, FITNESS
FOR A PARTICULAR PURPOSE AND NONINFRINGEMENT. IN NO EVENT SHALL THE AUTHORS OR
COPYRIGHT HOLDERS BE LIABLE FOR ANY CLAIM, DAMAGES OR OTHER LIABILITY, WHETHER
IN AN ACTION OF CONTRACT, TORT OR OTHERWISE, ARISING FROM, OUT OF OR IN
CONNECTION WITH THE SOFTWARE OR THE USE OR OTHER DEALINGS IN THE SOFTWARE.
\end{quote}

\section{Broader Impacts}
This work studies image tokenizers in the context of unified pure-AR multimodal models. By showing how tokenizer design affects modeling, including cross-modal modeling and text-side behavior, our analysis may help researchers design and diagnose multimodal training beyond reconstruction or single-axis evaluation. Improved tokenizer analysis could contribute to stronger generation and understanding models, which may support creative, educational, and accessibility applications. 

However, stronger multimodal models also raise risks such as biased outputs, misleading synthetic content, and misuse. Our experiments use publicly available images and do not introduce a deployed system, but future work using these insights should be paired with standard safety, bias, privacy, and responsible-release evaluations.

\end{document}